\documentclass[letterpaper]{article} 
\usepackage{aaai2026}  
\usepackage{times}  
\usepackage{helvet}  
\usepackage{courier}  
\usepackage[hyphens]{url}  
\usepackage{graphicx} 
\usepackage{natbib}  
\usepackage{caption} 
\usepackage{algorithm}
\usepackage{algorithmic}
\usepackage{booktabs} 
\usepackage{amsmath}
\usepackage{newfloat}
\usepackage{listings}
\DeclareCaptionStyle{ruled}{labelfont=normalfont,labelsep=colon,strut=off}
\floatstyle{ruled}
\newfloat{listing}{tb}{lst}{}
\floatname{listing}{Listing}

\title{SE360: Semantic Edit in 360° Panoramas via Hierarchical Data Construction}
\author{
    Haoyi Zhong\textsuperscript{\rm 1},
    Fang-Lue Zhang\textsuperscript{\rm 1}\thanks{Corresponding author},
    Andrew Chalmers\textsuperscript{\rm 1},
    Taehyun Rhee\textsuperscript{\rm 2}
}
\affiliations{
    \textsuperscript{\rm 1}Victoria University of Wellington, New Zealand\\
    \textsuperscript{\rm 2}The University of Melbourne, Australia\\
    \{haoyi.zhong, fanglue.zhang, andrew.chalmers\}@vuw.ac.nz, taehyun.rhee@unimelb.edu.au
}

\nocopyright
\begin{document}

\maketitle

\begin{abstract}
While instruction-based image editing is emerging, extending it to 360° panoramas introduces additional challenges. Existing methods often produce implausible results in both equirectangular projections (ERP) and perspective views. To address these limitations, we propose SE360, a novel framework for multi-condition guided object editing in 360° panoramas. At its core is a novel coarse-to-fine autonomous data generation pipeline without manual intervention. This pipeline leverages a Vision-Language Model (VLM) and adaptive projection adjustment for hierarchical analysis, ensuring the holistic segmentation of objects and their physical context. The resulting data pairs are both semantically meaningful and geometrically consistent, even when sourced from unlabeled panoramas. Furthermore, we introduce a cost-effective, two-stage data refinement strategy to improve data realism and mitigate model overfitting to erase artifacts. Based on the constructed dataset, we train a Transformer-based diffusion model to allow flexible object editing guided by text, mask, or reference image in 360° panoramas. Our experiments demonstrate that our method outperforms existing methods in both visual quality and semantic accuracy.
\end{abstract}

\begin{links}
    \link{Code}{https://github.com/zhonghaoyi/SE360.git}
\end{links}

\section{Introduction}

While instruction-based image editing has made substantial progress in conventional image settings, the growing demand for immersive applications—such as virtual reality (VR)—has shifted attention toward the 360° panoramic images. Enabling diverse, high-fidelity content manipulation in omnidirectional environments presents additional challenges that remain underexplored. 

\begin{figure}[t]
\centering
\includegraphics[width=1.00\columnwidth]{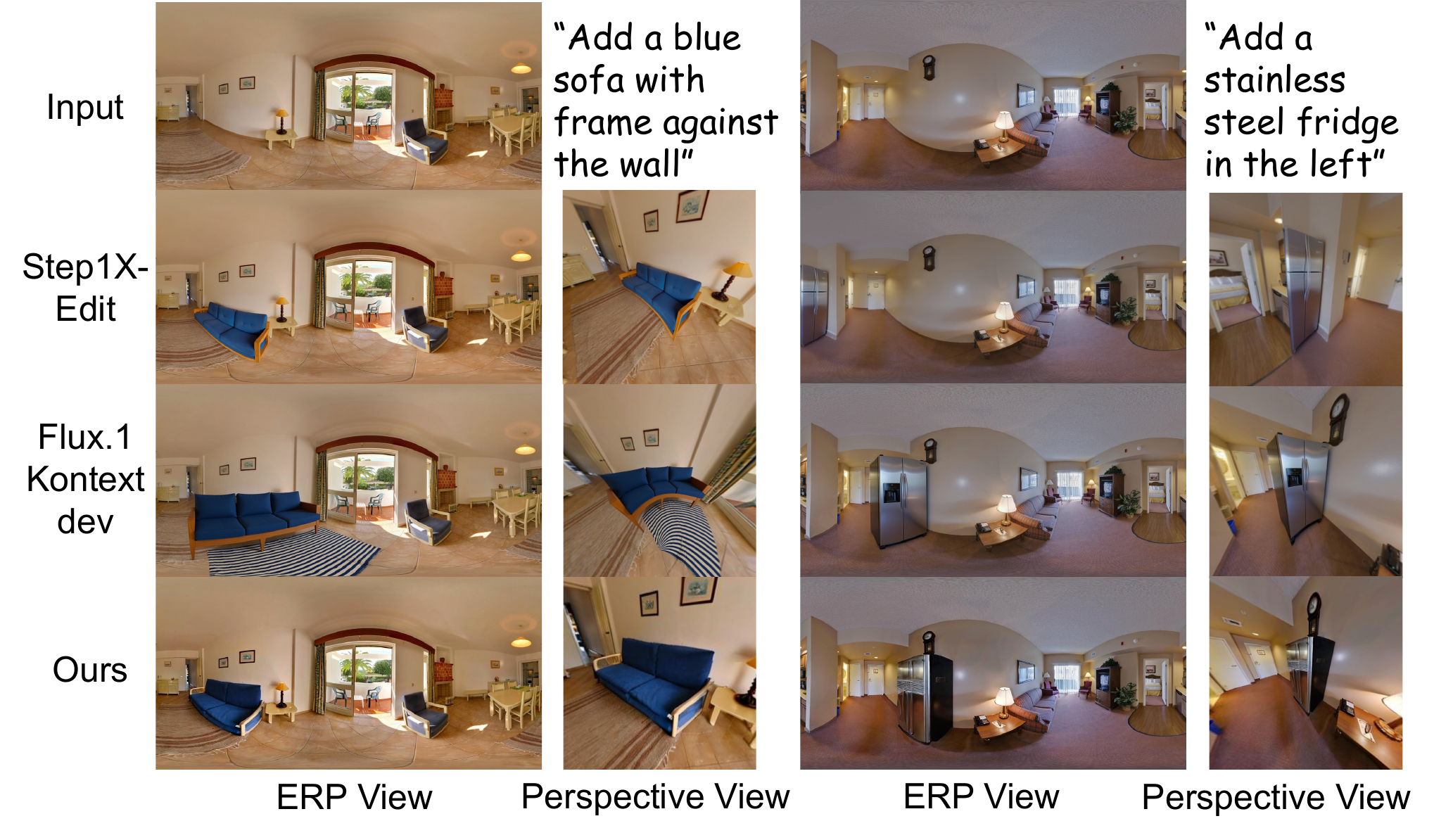}
\caption{Comparison with state-of-the-art models.}
\label{fig:pictures/Picture_1.pdf}
\end{figure}

State-of-the-art instruction-based image editing methods often produce artifacts when applied to 360° imagery in equirectangular projection (ERP). As shown in Figure~\ref{fig:pictures/Picture_1.pdf}, these models struggle with the domain gap between perspective and spherical views—edited objects appear plausible in ERP but become geometrically distorted when viewed through VR or panoramic viewers. Lacking awareness of spherical continuity, they also treat ERP boundaries as disconnected, causing visual breaks. Cubemap-based approaches like Omni$^2$\cite{omni2} offer partial relief by reducing distortion but yield moderate gains in realism.

However, Cubemap-based approaches introduce their own limitations, particularly in handling large objects that span multiple cubemap faces. More critically, automated data annotation methods based on Vision-Language Models (VLMs), such as Erasedraw ~\cite{erasedraw}, Omni$^2$ and InsightEdit~\cite{insightedit} often rely on bounding box (BBox)-guided segmentation for mask generation. These methods tend to overlook object hierarchies and physical containment relationships. For example, a sofa might be segmented while ignoring a cushion resting on it, resulting in semantically inconsistent training data. While such issues can be mitigated with manual filtering in perspective image datasets, this is impractical for the sparsely annotated 360° domain.

To address these challenges, we propose \textbf{SE360}, a scalable framework for instruction-guided object editing in 360° panoramas. At its core is a coarse-to-fine autonomous data generation pipeline that creates high-quality training pairs from unlabeled ERP images. The pipeline proceeds as follows: (1) \textit{Object identification}: a VLM, aided by referential detection models, extracts candidate objects and generates descriptions in the ERP domain. (2) \textit{Containment-aware projection}: candidate objects are projected into perspective views, where hierarchical relationships (e.g., sofa and cushion) are identified via VLM-driven analysis. (3) \textit{View adjustment and segmentation}: the system adjusts viewpoint and field of view (FOV) to fully capture the object and its context, followed by segmentation to obtain object masks. (4) \textit{Erasing and prompt synthesis}: the object is removed using the mask and an erasing model, while the VLM regenerates descriptive prompts based on the original attributes and spatial context. This process requires no manual labeling and enables large-scale, semantically aligned 360° editing data generation.

Beyond the base pipeline, we address limitations in erasing artifacts and model overfitting through refinement and training design. Using these two datasets, we train a multi-conditional Transformer diffusion model in a two-stage process to perform object editing. To enable 360° spatial understanding, we replace conventional positional embeddings with 3D spherical positional embedding that projects each pixel to unique coordinates on a unit sphere. This provides explicit geometric priors about distortion patterns and boundary continuity in panoramic imagery.

Our main contributions are summarized as follows:

\begin{itemize}
\item We propose a novel, coarse-to-fine automated 360° data generation pipeline. By leveraging adaptive projection adjustment and VLM-driven analysis, it achieves a holistic understanding of object hierarchies and addresses the challenge of editing large or cross-face objects.
\item We introduce a cost-effective data refinement strategy that combines traditional erasing with advanced instruction-based models, significantly improving training data quality.
\item We build and train a multi-conditional diffusion-based 360° panorama editing model. The model incorporates spherical positional priors and topological continuity handling, enabling instruction-guided editing across arbitrary locations.
\end{itemize}

Our experiments demonstrate that SE360 outperforms existing methods across multiple metrics, excelling in visual quality, semantic accuracy, and edit plausibility.

\section{Related Works}
\textbf{Instruction-Guided Image Editing.} Instruction-guided image editing aims to modify images via natural language instructions. The foundational paradigm for this task was established by the pioneering work InstructPix2Pix \cite{InstructPix2Pix}, which auto-generates large-scale (instruction, image) training data, surpassing earlier explorations like SDEdit \cite{sdedit,prompt2promopt}.
To enhance the quality of this synthetic data, subsequent works have pursued multiple avenues, including leveraging high-quality human annotations \cite{magicbrush} and proposing innovative automatic data generation strategies \cite{ultraedit, hqedit, erasedraw, paintbyinpaint, insightedit}. More recently, the advent of Diffusion Transformers \cite{dit} has propelled new paradigms like "in-context editing" and spurred research into building unified models for both generation and editing \cite{omnigen, unireal, icedit, fluxkontext, step1x}. However, all these methods face fundamental challenges when applied to 360° panoramas. As they are not designed to account for the unique properties of spherical geometry, their edits result in severe perspective distortions and visible seam artifacts. Our work is proposed to address this specific challenge.

\noindent\textbf{360° Panorama Generation and Editing.} 360° panorama editing has unique challenges related to continuity and geometric consistency for generative models. To ensure seamless connectivity between the left and right boundaries, existing methods either employ an autoregressive outpainting strategy to progressively expand from a narrow FoV image \cite{autoregressive, panodiffusion}, or leverage techniques such as circular padding and blending during the holistic generation process to enforce topological continuity on the model \cite{diffusion360, panfusion}. Some methods circumvent the distortion issues inherent in the ERP format by decomposing the panorama into multiple perspective views with lower distortion, such as cubemaps, thereby better leveraging powerful diffusion models pre-trained on standard perspective images \cite{mvdiffusion, cubediff}.
However, the task of editing panoramas, particularly at the semantic level, remains underexplored. Early panoramic editing techniques \cite{efficient, fast} allow simple color or style adjustments, lacking semantic content understanding. Approaches like \cite{omni2} have begun to unify generation and editing tasks; however, a dedicated framework for high-fidelity, instruction-guided, object-level editing—particularly one capable of handling objects that span across multiple views—remains an open challenge. Our work is proposed to bridge this gap to enable consistent semantic editing of arbitrary objects within 360° panoramas.

\section{Method}
SE360 employs a transformer-based learning framework capable of handling diverse input conditions, thereby enabling flexible and high-fidelity panoramic image editing. To train the model, we adopt a two-phase, automated data generation pipeline to construct 360° panoramic datasets. The initial phase, \textit{SE360-Base}, is designed to extract fine-grained, instance-level annotations from large-scale 360° panorama datasets. The subsequence phase, \textit{SE360-HF}, builds upon the output of the SE360-Base, refines the data via enhancing the images' visual quality. Finally, we obtain paired image data for the most common editing tasks: object removal and addition.

\subsection{SE360-Base: Large-Scale Dataset Generation}

Through several dedicated data processing stages, SE360-Base enables robust object grounding, captures complete composite objects, and ensures full visibility in 360° scenes. This results in high-quality, context-rich annotations that support accurate and flexible image editing.

\subsubsection{Stage 1: Object Extraction.}
As in Figure \ref{fig2}, we first guide the VLM model (Qwen2.5-VL-32B)~\cite{qw2.5vl} to identify and describe the primary foreground objects within each scene, yielding a structured list of objects, each annotated with a detailed description and a category.
Subsequently, we employ a multi-model fusion strategy for robust phrase grounding, which integrates the strengths of three distinct models: the initial VLM, Grounding DINO~\cite{groundingdino}, and Florence-2~\cite{florence2}. We adopt a consensus-based filtering strategy to enhance grounding precision. An object is accepted only if Grounding DINO or Florence-2 produces a bounding box with high Intersection over Union (IoU) overlap with the VLM’s initial detection, and the largest valid box is selected for coverage. For objects missed by the VLM but detected by both others, we retain only high-confidence overlapping results. Non-Maximum Suppression and geometric filters are then applied to remove redundant or noisy detections.

\subsubsection{Stage 2: Physical Affiliation Analysis.}
Recent methods~\cite{erasedraw, insightedit} using VLMs and Grounded-SAM~\cite{groundedsam} often produce incomplete masks, especially for composite objects with varied textures—e.g., segmenting a side table but missing items placed on it (Figure \ref{fig3}). Even replacing Segment Anything (SAM)~\cite{sam} with advanced models like Semantic-SAM~\cite{semanticsam} struggles to unify such objects into a single entity. To overcome this, we propose Physical Affiliation Analysis (PAA), which uses VLM-guided prompts to identify and list items physically associated with a target object. It enables SAM to segment each part individually, preserving object completeness for fine-grained detection. We automate this process using a VLM guided by a rule-based prompt. The VLM is prompted to list items that are on, inside, or attached to the target object. The results are saved in a structured JSON file for use in the next fine-grained detection stage.

\subsubsection{Stage 3: Adaptive Projection Adjustment.}
\begin{figure*}[t]
\centering
\includegraphics[width=2.00\columnwidth]{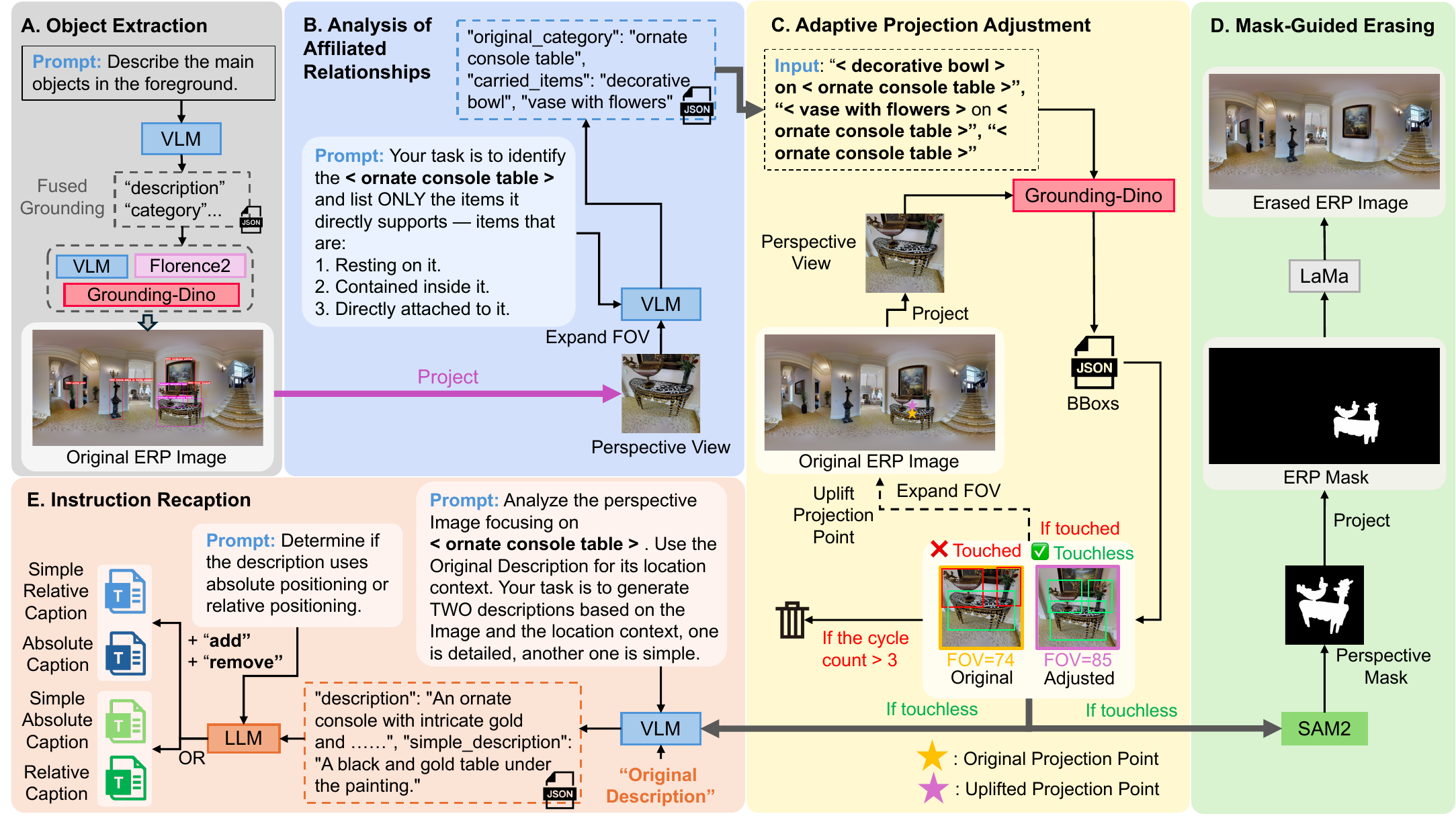} 

\caption{The overview diagram of SE360-Base.}
\label{fig2}
\end{figure*}

\begin{figure}[t]
\centering
\includegraphics[width=1.00\columnwidth]{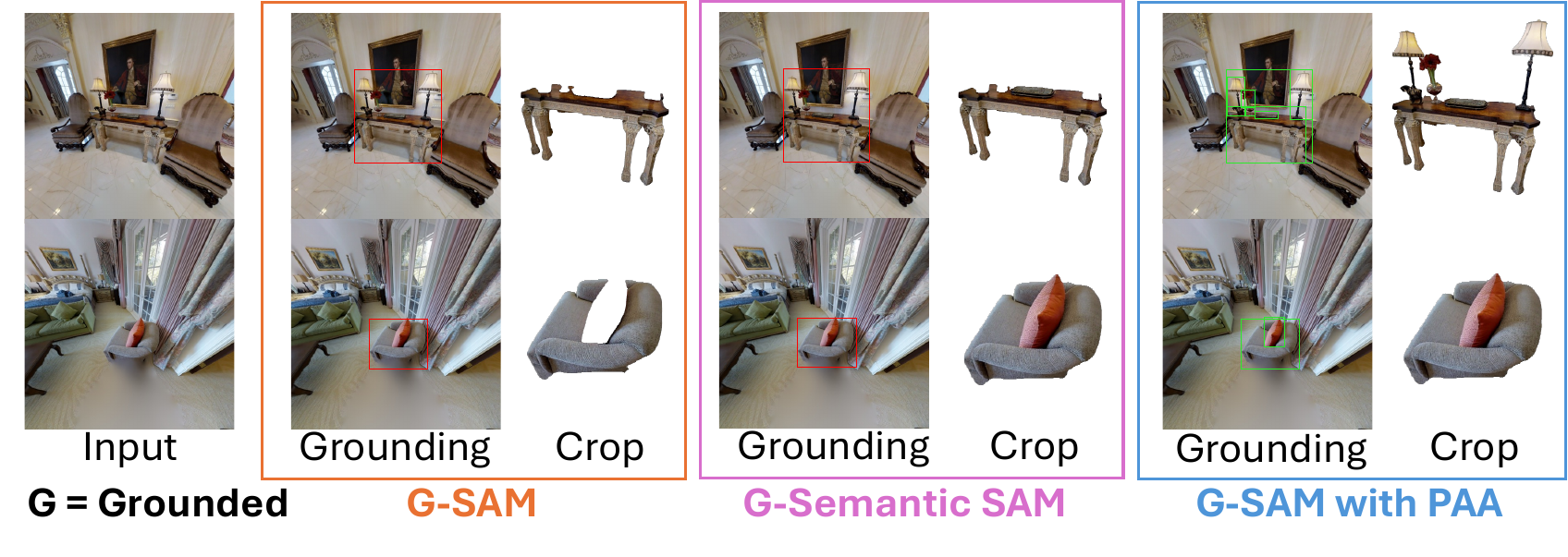} 

\caption{Comparison of different segmentation strategies.}
\label{fig3}
\end{figure}

We then perform fine-grained detection within an adaptively adjusted perspective viewport. Rather than relying on a static projection, our method iteratively adjusts the Field of View (FOV) and projection center to ensure all target objects are fully visible, as illustrated in the "Adaptive Adjustment" panel of Figure \ref{fig2}.
Grounding DINO is applied to detect primary and affiliated items in the initial projection. If any bounding box touches the viewport edge—indicating partial visibility—the FOV is expanded and the center refined. This loop continues until all objects are fully enclosed or a maximum number of iterations is reached. This adaptive mechanism ensures complete object views for reliable segmentation in complex 360° scenes.

\subsubsection{Stage 4: Mask-Guided Erasing.}
To support mask-guided erasing, we employ a progressive segmentation strategy with SAM2~\cite{sam2}. An initial coarse mask, generated by SAM2 using the object's bounding box, is refined by re-prompting SAM2 with positive points sampled from its interior. We then remove noises and dilate the refined mask to create the final erasing mask. Instead of erasing on the perspective image, we reproject the final mask to the full ERP image to guide LaMa~\cite{lama}, leveraging global context for more coherent results.

\subsubsection{Stage 5: Instruction Recaption.}
We finally introduce an Instruction Recaption module to enhance the descriptive richness of our dataset. Leveraging the original description and the adaptively-adjusted perspective view, we utilize a VLM to generate two new, hierarchical descriptions:
\begin{itemize}
\item Standard Refined Description, which integrates verifiable visual details from the high-quality perspective view with a pre-established spatial context.
\item Brief Description, which offers a succinct summary of the object's core attributes and location.
\end{itemize}
This dual-level description provides each object with labels that are both detailed and context-aware, as well as concise and easily parsable—allowing models to handle instructions of varying complexity. The resulting JSON file is then processed by Qwen3-8B~\cite{yang2025qwen3}, which classifies the localization method as either absolute (relative to the image frame) or relative (with respect to other objects). This classification guides our image rotation strategy during model training.
From the initial data generation phase, we obtained 195,504 editing triplets (37,304 × 2 + 60,448 × 2) for addition and removal tasks. This total includes both the detailed and brief variants of each instruction.

\subsection{SE360-HF: High-Fidelity Refinement}
\begin{figure}[t]
\centering
\includegraphics[width=1.00\columnwidth]{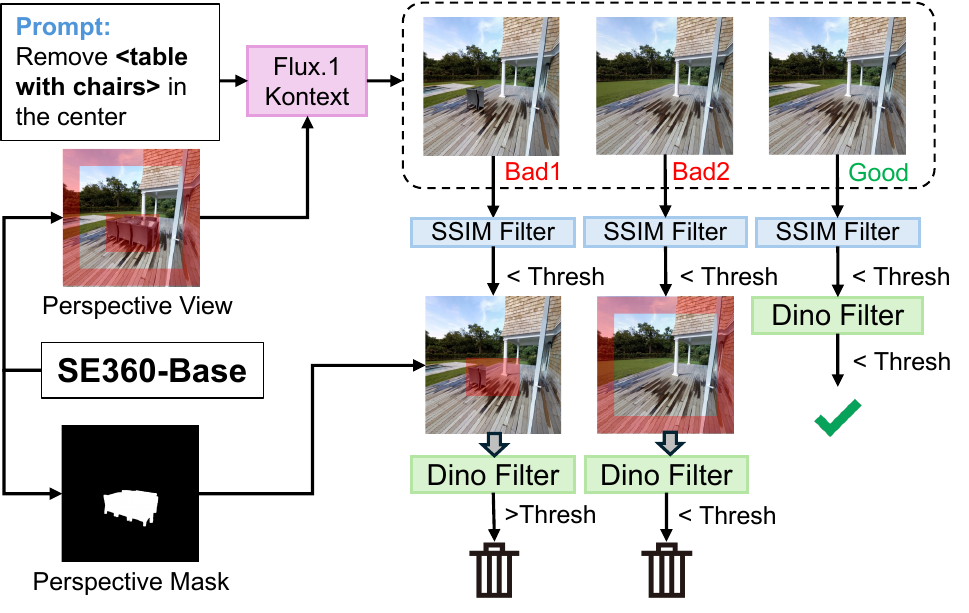} 

\caption{The overview diagram of SE360-HF.}
\label{fig4}
\end{figure}

Object erasing in complex scenes often results in visual artifacts, such as residual shadows, which can cause the model to overfit to these imperfections when generating novel content, as shown in our ablation studies in later sections. Although downsampling helps mitigate such artifacts during training, it proves unreliable for large-scale objects.
To address this, we employ a high-fidelity data generation phase, SE360-HF. We replace LaMa with a more robust, instruction-driven editor—Flux.1 Kontext max—to produce cleaner image-edit pairs. As shown in Figure \ref{fig4}, we first generate a template instruction based on the object's category (e.g., “remove the table with chairs in the center of the image”), which, along with the perspective view from the SE360-Base pipeline, guides the editing process. The resulting images undergo a two-stage filtering process:
\begin{enumerate}
    \item SSIM Filtering: We first compute the Structural Similarity Index (SSIM)~\cite{psnrssim} between the erased and original images, discarding those with scores above a predefined threshold.
    
    \item Region-Constrained Feature Similarity Check: We perform a localized similarity check using DINOv2~\cite{dinov2} features within the masked bounding box and at the image borders. This region-constrained strategy, unlike a global comparison, allows for consistent filtering across objects of varying sizes with a uniform threshold. Images with high similarity scores in these areas are excluded.
\end{enumerate}

We sampled 10,000 images with a distribution of 70\% large, 20\% medium, and 10\% small. After applying the automated filtering pipeline, 8,420 high-quality images were retained, resulting in 16,840 edit pairs using the same instructions as SE360-Base. Training on this refined dataset reduces overfitting to artifacts and improves the realism of generated objects.

\subsection{Model Design}
Here, we introduce SE360, our learning framework that adapts powerful pre-trained diffusion models for multi-conditional, multi-modal editing of 360° panoramic images through a series of designs tailored to the spherical domain.

\subsubsection{Diffusion Transformer.}
\begin{figure*}[t]
\centering
\includegraphics[width=2.00\columnwidth]{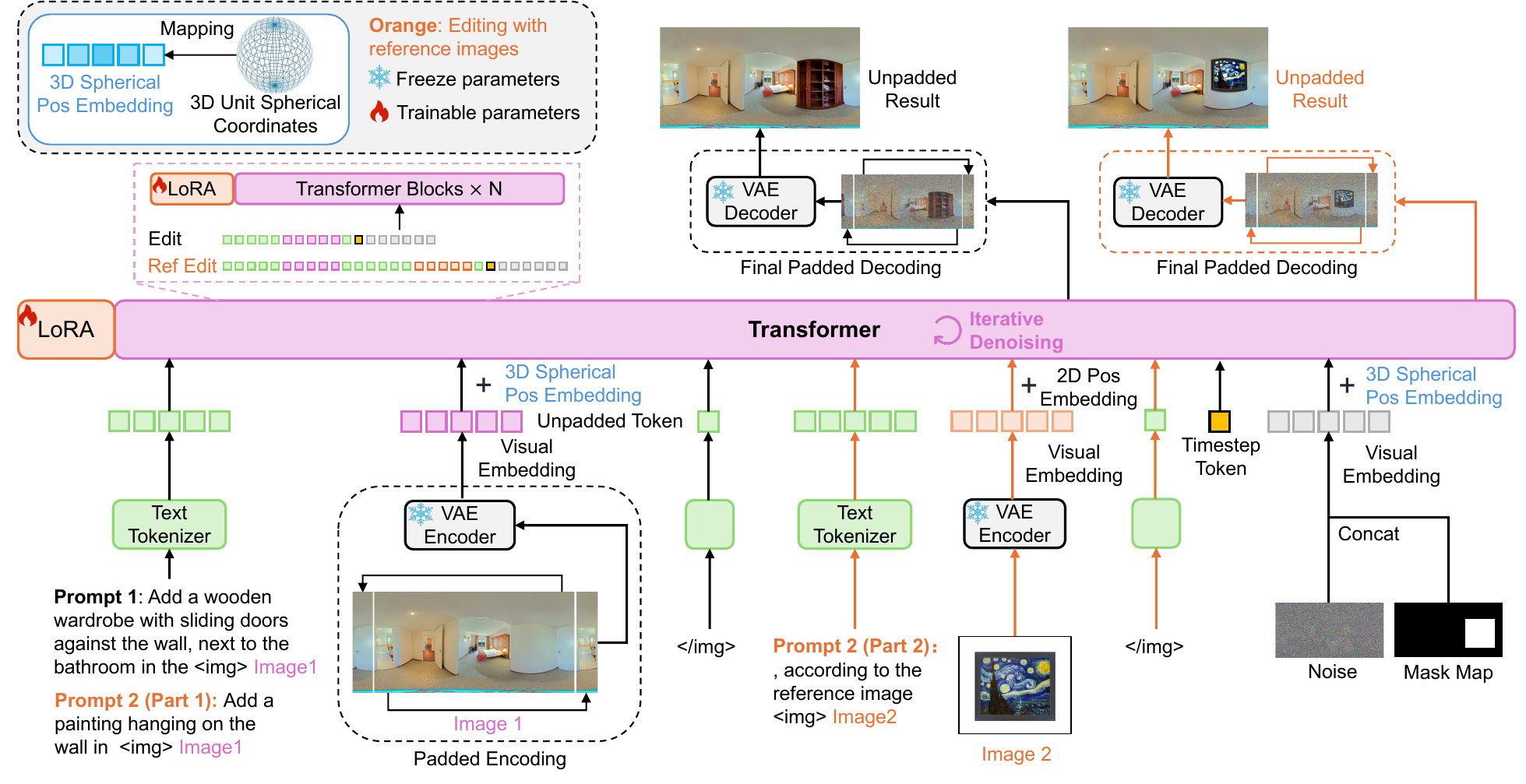} 

\caption{The model architecture of SE360.}
\label{fig5}
\end{figure*}

As illustrated in Figure~\ref{fig5}, the core of our model is a Diffusion Transformer that drives the generative process. To incorporate strong editing priors, we initialize it with weights from the pre-trained OmniGen~\cite{omnigen}. The Transformer operates on a concatenated 1D sequence that fuses multimodal information, including token embeddings of text prompts and latent visual features extracted from the source panorama and reference images—produced by a frozen, location-aware Variational Autoencoder (VAE)~\cite{vae} of SDXL~\cite{sdxl} . This sequence is further enhanced with 3D spherical positional embeddings for spatial awareness, along with timestep embeddings and an editing mask map to guide the diffusion.

For parameter-efficient adaptation to panoramic editing, we adopt Low-Rank Adaptation (LoRA)~\cite{lora}, fine-tuning only the query (Q), key (K), and value (V) projection matrices within attention layers, as well as the feed-forward networks. This strategy preserves most of the pre-trained knowledge while efficiently adapting the model to the unique properties of the 360° spherical domain.

\subsubsection{Location-Aware Encoding and Decoding.}
To eliminate visible artifacts at the longitudinal seams of panoramic images caused by standard VAE convolutional kernels failing to account for spherical topology~\cite{lpo}, we introduce a symmetric "Pad-and-Unpad" strategy. During encoding, we apply circular padding by prepending and appending content from the opposite horizontal edges, providing the encoder with adjacent 360° context. This results in a semantically seamless latent representation, which is then cropped back to the original dimensions. In decoding, we apply the same circular padding to the latent features, allowing the decoder to reconstruct smooth boundary transitions at the pixel level. Finally, we crop the central portion of the slightly wider output to produce a seamless and visually coherent panoramic image, effectively removing artifacts introduced by standard convolution.

\subsubsection{3D Spherical Positional Embedding.}
Conventional 2D sinusoidal positional embeddings struggle with panoramas, as they fail to capture the true spatial relationships inherent in spherical geometry. To overcome this, we introduce a 3D Spherical Positional Embedding (SPE) strategy. Specifically, we map ERP pixel coordinates to 3D Cartesian coordinates on a unit sphere. Given pixel coordinates $(i, j)$ in an ERP image of size $H \times W$, where $i \in [0, H{-}1]$ and $j \in [0, W{-}1]$, we first convert them to spherical coordinates—longitude ($\lambda$) and latitude ($\phi$)—and then transform them into 3D Cartesian coordinates $\mathbf{p}$ on the unit sphere as:
\begin{equation}
\mathbf{p} = (x, y, z) = (\cos\phi\cos\lambda, \cos\phi\sin\lambda, \sin\phi).
\label{eq:cartesian}
\end{equation}

We extend the standard frequency-based positional encoding to 3D by dividing the embedding dimension $d$ equally among the $x$, $y$, and $z$ axes, applying sinusoidal functions independently to each. This preserves spatial proximity on the spherical manifold while remaining compatible with Transformer architectures, offering a more faithful spatial prior for panoramic understanding.

\subsubsection{Training Methodology.}
SE360 contains 3.9 billion parameters and is initialized from OmniGen. We adopt a two-stage training strategy: first, foundational training on SE360-Base for 10K iterations; second, fine-tuning on SE360-HF for 1,000 iterations, using weights from stage one. Training is based on the Flow Matching~\cite{flowmaching} framework and incorporates LoRA with rank and alpha set to 16. To enable precise local editing, a mask map is applied with 20\% probability. For reference-based addition, we use object masks from perspective views in SE360-Base to remove backgrounds and construct inputs. We also conduct orientation-decoupled data augmentation by latitudinal rotation to preserve directional consistency with the text prompt containing absolute instructions. Further details are provided in the supplementary. Data augmentation, including scaling and rotation, is applied to enhance robustness.

\section{Experiments}
We conduct a series of experiments to comprehensively evaluate SE360. Our evaluation focuses on three aspects: comparison with state-of-the-art image editing models for panoramic object removal and addition, demonstration of multi-conditional input editing, and ablation studies analyzing each data generation pipeline’s contribution.

\begin{figure*}[t]
\centering
\includegraphics[width=2.00\columnwidth]{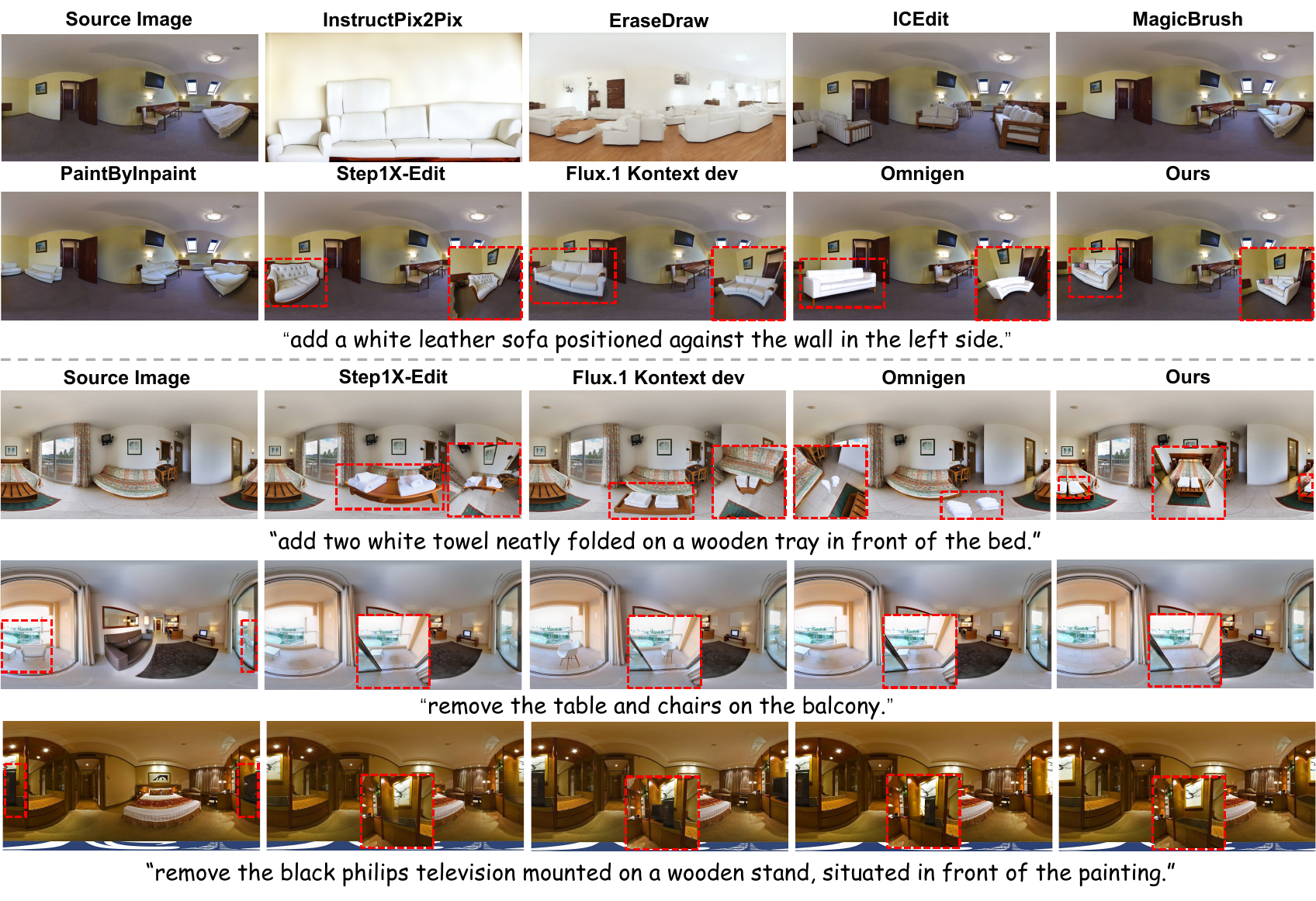} 

\caption{The comparison with SOTA models.}
\label{fig6}
\end{figure*}

\subsection{Dataset}
SE360 builds a large-scale editing dataset of over 30,000 images from Matterport3D~\cite{matterport3d} and Structured3D~\cite{structured3d}. 
For evaluation, we curated a high-quality test set from the PanoContext~\cite{panocontext} dataset, named by \textit{PanoEval}. We randomly selected 250 panoramas and applied the same data creation pipeline as SE360-HF to generate editing pairs. These pairs were then manually inspected to ensure correctness and quality, resulting in a test set of 448 high-fidelity editing pairs.

\subsection{Comparison with State-of-the-art}
To evaluate the model's handling of panoramic boundaries, we created a challenging subset, PanoEval-Boundary, containing 200 editing pairs from our test set that were rotated horizontally so the object of interest lies at or crosses the left–right seam of the ERP image. The main text reports only average results across the two test sets; detailed results are provided in the supplementary material.

To evaluate the performance of SE360 in 360° panoramic image editing, we benchmark it against several state-of-the-art open-source perspective image editing models, including InstructPix2Pix~\cite{InstructPix2Pix}, Erasedraw~\cite{erasedraw}, ICEdit~\cite{icedit}, MagicBrush~\cite{magicbrush}, PaintByInpaint~\cite{paintbyinpaint}, Omnigen~\cite{omnigen}, Flux.1 Kontext dev~\cite{fluxkontext} and Step1X-Edit~\cite{step1x}, on two primary tasks: object removal and object addition.

\begin{table}[t]
\centering

\small
\setlength{\tabcolsep}{5pt} 
\begin{tabular}{@{}lccc@{}} 
\toprule
Model & CS$_{\text{pers}}$$\uparrow$ & LPIPS$\downarrow$ & LPIPS$_{\text{pers}}$$\downarrow$ \\ 
\midrule
InstructPix2Pix & 23.691 & 0.783 & 0.798 \\
Erasedraw & 25.262 & 0.733 & 0.737 \\
ICEdit & 27.196 & 0.286 & 0.356 \\
MagicBrush & 27.618 & 0.109 & 0.219 \\
PaintByInpaint & 28.124 & 0.094 & 0.198 \\
OmniGen & 28.370 & 0.091 & 0.209 \\
Flux.1 Kontext dev & 29.295 & 0.141 & 0.265 \\
Step1X-Edit & 28.919 & 0.091 & 0.211 \\
\midrule
\textbf{SE360 (Ours)} & \textbf{29.354} & \textbf{0.076} & \textbf{0.177} \\
\bottomrule
\end{tabular}

\caption{Quantitative comparison for the object addition task on our test set. Best results are in \textbf{bold}.}
\label{tab:addition}
\end{table}

\begin{table*}[t]
\centering

\small
\begin{tabular}{l|ccc|ccccc}
\toprule
& \multicolumn{3}{c|}{\textbf{Addition}} & \multicolumn{5}{c}{\textbf{Removal}} \\
\midrule
Trainnig Type & CS$_{\text{pers}}$$\uparrow$ & LPIPS$\downarrow$ & LPIPS$_{\text{pers}}$$\downarrow$ & CS-No$_{\text{pers}}$$\uparrow$ & FAED$\downarrow$ & LPIPS$\downarrow$ & LPIPS$_{\text{pers}}$$\downarrow$ & PSNR$\uparrow$ \\
\midrule
SE360-Base w/artifacts & \textbf{29.377} & \textbf{0.053} & \textbf{0.125} & - & - & - & - & - \\
\midrule
SE360-Base & 27.965 & \textbf{0.074} & 0.184 & 73.055 & 0.498 & \textbf{0.057} & \textbf{0.128} & 25.854 \\
SE360-HF & 28.402 & 0.171 & 0.285 & \textbf{73.820} & 2.570 & 0.170 & 0.268 & 19.419 \\
Full & \textbf{29.354} & 0.076 & \textbf{0.177} & 73.143 & \textbf{0.314} & 0.079 & 0.179 & \textbf{26.394} \\
\bottomrule
\end{tabular}

\caption{Ablation studies on the object addition and removal tasks.}
\label{tab:ablation_insert_remove}
\end{table*}

\begin{table}[t]
\centering

\small
\setlength{\tabcolsep}{0.7pt} 
\begin{tabular}{l ccccc}
\toprule
Model & CS-No$_{\text{pers}}$$\uparrow$ & FAED$\downarrow$ & LPIPS$\downarrow$ & LPIPS$_{\text{pers}}$$\downarrow$ & PSNR$\uparrow$ \\
\midrule
MagicBrush & 71.319 & 2.798 & 0.119 & 0.223 & 20.268 \\
ICEdit & 72.232 & 1.545 & 0.268 & 0.311 & 23.058 \\
Omnigen & 72.256 & 0.857 & 0.066 & 0.142 & 25.200 \\
Flux.1 Kontext dev & 71.975 & 1.580 & 0.115 & 0.191 & 19.890 \\
Step1X-Edit & 72.606 & 0.748 & \textbf{0.057} & \textbf{0.095} & \textbf{28.471} \\
\midrule
\textbf{SE360 (Ours)} & \textbf{73.143} & \textbf{0.314} & 0.079 & 0.179 & 26.394 \\
\bottomrule
\end{tabular}

\caption{Quantitative comparison for the object removal task on our test set. Best results are in \textbf{bold}.}
\label{tab:removal}
\end{table}

\subsubsection{Evaluation Metrics.}
We use PSNR~\cite{psnrssim} and LPIPS~\cite{lpips} to measure reconstruction quality and perceptual similarity in panoramas, respectively. LPIPS$_{\text{pers}}$ are used to measure perceptual similarity in perspective view. FAED~\cite{faed} assesses the realism of the generated panoramas. CS$_{\text{pers}}$ measures the semantic consistency between the edited result and the text description by CLIP ~\cite{clip} in perspective views. For the removal task, we utilize CS-No$_{\text{pers}}$, which measures the similarity of the edited image to a scene description that excludes the target object in perspective view. The pespective view projected from predefined viewpoints with a FOV of at least 100°.

\subsubsection{Results.} SE360 outperforms existing methods in object addition and removal (Tables \ref{tab:addition} and \ref{tab:removal}), achieving superior perceptual quality and semantic accuracy. These advantages are shown in Figure \ref{fig6}. Other models often produce geometrically inconsistent objects or leave visual artifacts when handling large boundary-crossing edits. In contrast, SE360 leverages its 3D spatial understanding to generate perspectively accurate objects, seamlessly inpaint occluded regions, and produce visually coherent results. For object removal across stitching seams, methods like Step1X-Edit often fail to synthesize continuous textures, whereas SE360 removes objects cleanly and generates plausible background content. For object addition, especially with perspective distortion (e.g., inserting a painting high on a wall), existing methods tend to produce flat, unrealistic results, while SE360 generates properly warped, well‑lit additions.

\subsection{Multi-Condition Editing}
\begin{figure}[t]
\centering
\includegraphics[width=1.00\columnwidth]{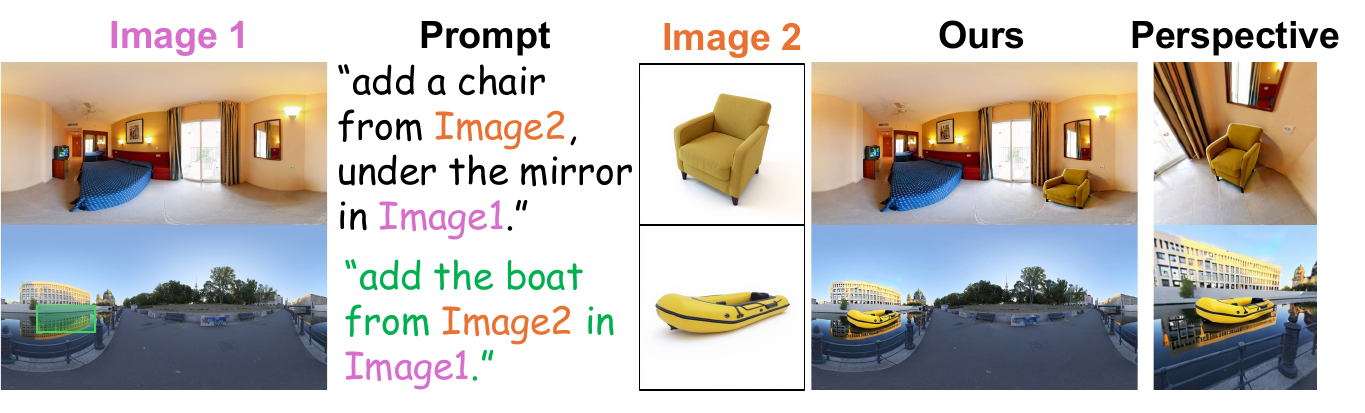} 

\caption{Addition with reference image and mask map.}
\label{fig7}
\end{figure}

Beyond single-object manipulation, SE360 proves capable of processing complex directives with multiple concurrent conditions. As showcased in Figure~\ref{fig7}, the model performs object insertion guided by both a mask map and a reference image, generating panoramas with geometric and photometric consistency.

\subsection{Ablation Studies}
To validate the effectiveness of our training method, we conducted a series of ablation studies on two test datasets. The average results in Table \ref{tab:ablation_insert_remove} reveal a clear trade-off in SE360's performance: the model performs best on data containing LaMa's erasing artifacts but worst when those artifacts are absent. This indicates that training exclusively on SE360-Base leads to overfitting to residual artifacts, even though the model learns object-level texture and geometry. Conversely, training solely on SE360-HF mitigates this overfitting but degrades generation quality due to insufficient training data. Fine-tuning a model pre-trained on SE360-Base with SE360-HF alleviates artifact overfitting while preserving the model's core generative abilities.


\begin{table}[t]
\centering
\small 
\setlength{\tabcolsep}{2.5pt} 
\begin{tabular}{lccc}
\toprule
Model &  Consistency $\uparrow$ & Quality $\uparrow$ &  Plausibility $\uparrow$ \\
\midrule
PaintByInpaint     & 2.04 (14.5\%) & 2.34 (15.6\%) & 2.44 (16.1\%) \\
Omnigen            & 2.11 (15.0\%) & 2.19 (14.6\%) & 2.34 (15.4\%) \\
Flux.1 Kontext dev & 2.91 (20.7\%) & 3.44 (22.9\%) & 3.11 (20.5\%) \\
Step1X-Edit        & 2.50 (17.8\%) & 2.94 (19.6\%) & 2.90 (19.1\%) \\
\midrule
\textbf{SE360 (Ours)} & \textbf{4.47 (31.9\%)} & \textbf{4.09 (27.3\%)} & \textbf{4.35 (28.7\%)} \\
\bottomrule
\end{tabular}

\caption{User Study results for object addition.}
\label{tab:user_study_very_compact}
\end{table}

\subsection{User Study}
We conducted a comprehensive user study to evaluate our method on the task of instruction-based object addition. In the study, 22 participants compared results from SE360 against four top-performing models across 30 distinct samples. Evaluations were performed on a 5-point Likert scale, assessing three key metrics: Consistency (alignment between the generated object and the instruction), Quality (visual fidelity of the added object), and Plausibility (spatial and perspective coherence). As in Table~\ref{tab:user_study_very_compact}, SE360 outperforms all competing methods. These results confirm that users perceive our generated objects as more faithful to the given instructions and visually superior. 


\section{Conclusion}
We presented SE360, a semantic editing framework for 360° panoramas that resolves plausible and discontinuity issues from existing methods. Its core is a coarse-to-fine data pipeline ensuring object integrity and realism via hierarchical analysis and two-stage refinement, which also mitigates artifact overfitting. The resulting high-quality dataset enables training of a geometry-aware Transformer diffusion model. Experimental results and user studies demonstrate that SE360 outperforms  prior approaches in visual quality, semantic accuracy, and geometric plausibility.
\paragraph{Limitations.}
Despite impressive performance, our method still has limitations. First, in reference-based tasks, the output of SE360 may bring background elements from the reference image to the result. Second, while our data generation pipeline incorporates multi-stage filtering, it may still introduce occasional errors.

\section*{Acknowledgments}
This work was supported by the Marsden Fund Council managed by the Royal Society of New Zealand (No. MFP-20-VUW-180).

\bibliography{aaai2026}

\begin{thebibliography}{46}
\providecommand{\natexlab}[1]{#1}

\bibitem[{Bai et~al.(2025)Bai, Chen, Liu, Wang, Ge, Song, Dang, Wang, Wang, Tang et~al.}]{qw2.5vl}
Bai, S.; Chen, K.; Liu, X.; Wang, J.; Ge, W.; Song, S.; Dang, K.; Wang, P.; Wang, S.; Tang, J.; et~al. 2025.
\newblock Qwen2. 5-vl technical report.
\newblock \emph{arXiv preprint arXiv:2502.13923}.

\bibitem[{Brooks, Holynski, and Efros(2023)}]{InstructPix2Pix}
Brooks, T.; Holynski, A.; and Efros, A.~A. 2023.
\newblock InstructPix2Pix: Learning to Follow Image Editing Instructions.
\newblock In \emph{{IEEE/CVF} Conference on Computer Vision and Pattern Recognition, {CVPR} 2023, Vancouver, BC, Canada, June 17-24, 2023}, 18392--18402.

\bibitem[{Canberk et~al.(2024)Canberk, Bondarenko, Ozguroglu, Liu, and Vondrick}]{erasedraw}
Canberk, A.; Bondarenko, M.; Ozguroglu, E.; Liu, R.; and Vondrick, C. 2024.
\newblock Erasedraw: Learning to insert objects by erasing them from images.
\newblock In \emph{European Conference on Computer Vision}, 144--160. Springer.

\bibitem[{Chang et~al.(2017)Chang, Dai, Funkhouser, Halber, Niessner, Savva, Song, Zeng, and Zhang}]{matterport3d}
Chang, A.; Dai, A.; Funkhouser, T.; Halber, M.; Niessner, M.; Savva, M.; Song, S.; Zeng, A.; and Zhang, Y. 2017.
\newblock Matterport3d: Learning from rgb-d data in indoor environments.
\newblock \emph{arXiv preprint arXiv:1709.06158}.

\bibitem[{Chen et~al.(2025)Chen, Zhang, Zhang, Zhou, Kim, Liu, Li, Zhang, Zhao, Wang et~al.}]{unireal}
Chen, X.; Zhang, Z.; Zhang, H.; Zhou, Y.; Kim, S.~Y.; Liu, Q.; Li, Y.; Zhang, J.; Zhao, N.; Wang, Y.; et~al. 2025.
\newblock Unireal: Universal image generation and editing via learning real-world dynamics.
\newblock In \emph{Proceedings of the Computer Vision and Pattern Recognition Conference}, 12501--12511.

\bibitem[{Feng et~al.(2023)Feng, Liu, Cui, and Xie}]{diffusion360}
Feng, M.; Liu, J.; Cui, M.; and Xie, X. 2023.
\newblock Diffusion360: Seamless 360 Degree Panoramic Image Generation based on Diffusion Models.
\newblock \emph{arXiv preprint arXiv:2311.13141}.

\bibitem[{Hertz et~al.(2023)Hertz, Mokady, Tenenbaum, Aberman, Pritch, and Cohen{-}Or}]{prompt2promopt}
Hertz, A.; Mokady, R.; Tenenbaum, J.; Aberman, K.; Pritch, Y.; and Cohen{-}Or, D. 2023.
\newblock Prompt-to-Prompt Image Editing with Cross-Attention Control.
\newblock In \emph{The Eleventh International Conference on Learning Representations, {ICLR} 2023, Kigali, Rwanda, May 1-5, 2023}.

\bibitem[{Hore and Ziou(2010)}]{psnrssim}
Hore, A.; and Ziou, D. 2010.
\newblock Image quality metrics: PSNR vs. SSIM.
\newblock In \emph{2010 20th international conference on pattern recognition}, 2366--2369.

\bibitem[{Hu et~al.(2021)Hu, Shen, Wallis, Allen{-}Zhu, Li, Wang, and Chen}]{lora}
Hu, E.~J.; Shen, Y.; Wallis, P.; Allen{-}Zhu, Z.; Li, Y.; Wang, S.; and Chen, W. 2021.
\newblock LoRA: Low-Rank Adaptation of Large Language Models.
\newblock \emph{arXiv preprint arXiv:2106.09685}.

\bibitem[{Hui et~al.(2024)Hui, Yang, Zhao, Shi, Wang, Wang, Zhou, and Xie}]{hqedit}
Hui, M.; Yang, S.; Zhao, B.; Shi, Y.; Wang, H.; Wang, P.; Zhou, Y.; and Xie, C. 2024.
\newblock Hq-edit: A high-quality dataset for instruction-based image editing.
\newblock \emph{arXiv preprint arXiv:2404.09990}.

\bibitem[{Kalischek et~al.(2025)Kalischek, Oechsle, Manhardt, Henzler, Schindler, and Tombari}]{cubediff}
Kalischek, N.; Oechsle, M.; Manhardt, F.; Henzler, P.; Schindler, K.; and Tombari, F. 2025.
\newblock Cubediff: Repurposing diffusion-based image models for panorama generation.
\newblock In \emph{The Thirteenth International Conference on Learning Representations}.

\bibitem[{Kingma and Welling(2014)}]{vae}
Kingma, D.~P.; and Welling, M. 2014.
\newblock Auto-Encoding Variational Bayes.
\newblock In \emph{2nd International Conference on Learning Representations, {ICLR} 2014, Banff, AB, Canada, April 14-16, 2014, Conference Track Proceedings}.

\bibitem[{Kirillov et~al.(2023)Kirillov, Mintun, Ravi, Mao, Rolland, Gustafson, Xiao, Whitehead, Berg, Lo et~al.}]{sam}
Kirillov, A.; Mintun, E.; Ravi, N.; Mao, H.; Rolland, C.; Gustafson, L.; Xiao, T.; Whitehead, S.; Berg, A.~C.; Lo, W.-Y.; et~al. 2023.
\newblock Segment anything.
\newblock In \emph{Proceedings of the IEEE/CVF international conference on computer vision}, 4015--4026.

\bibitem[{Labs et~al.(2025)Labs, Batifol, Blattmann, Boesel, Consul, Diagne, Dockhorn, English, English, Esser et~al.}]{fluxkontext}
Labs, B.~F.; Batifol, S.; Blattmann, A.; Boesel, F.; Consul, S.; Diagne, C.; Dockhorn, T.; English, J.; English, Z.; Esser, P.; et~al. 2025.
\newblock FLUX. 1 Kontext: Flow Matching for In-Context Image Generation and Editing in Latent Space.
\newblock \emph{arXiv preprint arXiv:2506.15742}.

\bibitem[{Li et~al.(2023)Li, Zhang, Sun, Zou, Liu, Yang, Li, Zhang, and Gao}]{semanticsam}
Li, F.; Zhang, H.; Sun, P.; Zou, X.; Liu, S.; Yang, J.; Li, C.; Zhang, L.; and Gao, J. 2023.
\newblock Semantic-SAM: Segment and Recognize Anything at Any Granularity.
\newblock \emph{arXiv preprint arXiv:2307.04767}.

\bibitem[{Lipman et~al.(2023)Lipman, Chen, Ben{-}Hamu, Nickel, and Le}]{flowmaching}
Lipman, Y.; Chen, R. T.~Q.; Ben{-}Hamu, H.; Nickel, M.; and Le, M. 2023.
\newblock Flow Matching for Generative Modeling.
\newblock In \emph{The Eleventh International Conference on Learning Representations, {ICLR} 2023, Kigali, Rwanda, May 1-5, 2023}.

\bibitem[{Liu et~al.(2025)Liu, Han, Xing, Yin, Wang, Cheng, Liao, Wang, Fu, Han et~al.}]{step1x}
Liu, S.; Han, Y.; Xing, P.; Yin, F.; Wang, R.; Cheng, W.; Liao, J.; Wang, Y.; Fu, H.; Han, C.; et~al. 2025.
\newblock Step1x-edit: A practical framework for general image editing.
\newblock \emph{arXiv preprint arXiv:2504.17761}.

\bibitem[{Liu et~al.(2024)Liu, Zeng, Ren, Li, Zhang, Yang, Jiang, Li, Yang, Su, Zhu, and Zhang}]{groundingdino}
Liu, S.; Zeng, Z.; Ren, T.; Li, F.; Zhang, H.; Yang, J.; Jiang, Q.; Li, C.; Yang, J.; Su, H.; Zhu, J.; and Zhang, L. 2024.
\newblock Grounding {DINO:} Marrying {DINO} with Grounded Pre-training for Open-Set Object Detection.
\newblock In \emph{Computer Vision - {ECCV} 2024 - 18th European Conference, Milan, Italy, September 29-October 4, 2024, Proceedings, Part {XLVII}}, volume 15105 of \emph{Lecture Notes in Computer Science}, 38--55.

\bibitem[{Lu et~al.(2024)Lu, Hu, Wang, Bai, and Wang}]{autoregressive}
Lu, Z.; Hu, K.; Wang, C.; Bai, L.; and Wang, Z. 2024.
\newblock Autoregressive omni-aware outpainting for open-vocabulary 360-degree image generation.
\newblock In \emph{Proceedings of the AAAI Conference on Artificial Intelligence}, volume~38, 14211--14219.

\bibitem[{Meng et~al.(2022)Meng, He, Song, Song, Wu, Zhu, and Ermon}]{sdedit}
Meng, C.; He, Y.; Song, Y.; Song, J.; Wu, J.; Zhu, J.; and Ermon, S. 2022.
\newblock SDEdit: Guided Image Synthesis and Editing with Stochastic Differential Equations.
\newblock In \emph{The Tenth International Conference on Learning Representations, {ICLR} 2022, Virtual Event, April 25-29, 2022}.

\bibitem[{Oh et~al.(2022)Oh, Cho, Chae, Park, Wang, and Yoon}]{faed}
Oh, C.; Cho, W.; Chae, Y.; Park, D.; Wang, L.; and Yoon, K.-J. 2022.
\newblock Bips: Bi-modal indoor panorama synthesis via residual depth-aided adversarial learning.
\newblock In \emph{European Conference on Computer Vision}, 352--371.

\bibitem[{Oquab et~al.(2023)Oquab, Darcet, Moutakanni, Vo, Szafraniec, Khalidov, Fernandez, Haziza, Massa, El-Nouby et~al.}]{dinov2}
Oquab, M.; Darcet, T.; Moutakanni, T.; Vo, H.; Szafraniec, M.; Khalidov, V.; Fernandez, P.; Haziza, D.; Massa, F.; El-Nouby, A.; et~al. 2023.
\newblock Dinov2: Learning robust visual features without supervision.
\newblock \emph{arXiv preprint arXiv:2304.07193}.

\bibitem[{Peebles and Xie(2023)}]{dit}
Peebles, W.; and Xie, S. 2023.
\newblock Scalable Diffusion Models with Transformers.
\newblock In \emph{{IEEE/CVF} International Conference on Computer Vision, {ICCV} 2023, Paris, France, October 1-6, 2023}, 4172--4182.

\bibitem[{Podell et~al.(2024)Podell, English, Lacey, Blattmann, Dockhorn, M{\"{u}}ller, Penna, and Rombach}]{sdxl}
Podell, D.; English, Z.; Lacey, K.; Blattmann, A.; Dockhorn, T.; M{\"{u}}ller, J.; Penna, J.; and Rombach, R. 2024.
\newblock {SDXL:} Improving Latent Diffusion Models for High-Resolution Image Synthesis.
\newblock In \emph{The Twelfth International Conference on Learning Representations, {ICLR} 2024, Vienna, Austria, May 7-11, 2024}.

\bibitem[{Radford et~al.(2021)Radford, Kim, Hallacy, Ramesh, Goh, Agarwal, Sastry, Askell, Mishkin, Clark et~al.}]{clip}
Radford, A.; Kim, J.~W.; Hallacy, C.; Ramesh, A.; Goh, G.; Agarwal, S.; Sastry, G.; Askell, A.; Mishkin, P.; Clark, J.; et~al. 2021.
\newblock Learning transferable visual models from natural language supervision.
\newblock In \emph{International conference on machine learning}, 8748--8763.

\bibitem[{Ravi et~al.(2024)Ravi, Gabeur, Hu, Hu, Ryali, Ma, Khedr, R{\"a}dle, Rolland, Gustafson et~al.}]{sam2}
Ravi, N.; Gabeur, V.; Hu, Y.-T.; Hu, R.; Ryali, C.; Ma, T.; Khedr, H.; R{\"a}dle, R.; Rolland, C.; Gustafson, L.; et~al. 2024.
\newblock Sam 2: Segment anything in images and videos.
\newblock \emph{arXiv preprint arXiv:2408.00714}.

\bibitem[{Ren et~al.(2024)Ren, Liu, Zeng, Lin, Li, Cao, Chen, Huang, Chen, Yan et~al.}]{groundedsam}
Ren, T.; Liu, S.; Zeng, A.; Lin, J.; Li, K.; Cao, H.; Chen, J.; Huang, X.; Chen, Y.; Yan, F.; et~al. 2024.
\newblock Grounded sam: Assembling open-world models for diverse visual tasks.
\newblock \emph{arXiv preprint arXiv:2401.14159}.

\bibitem[{Suvorov et~al.(2022)Suvorov, Logacheva, Mashikhin, Remizova, Ashukha, Silvestrov, Kong, Goka, Park, and Lempitsky}]{lama}
Suvorov, R.; Logacheva, E.; Mashikhin, A.; Remizova, A.; Ashukha, A.; Silvestrov, A.; Kong, N.; Goka, H.; Park, K.; and Lempitsky, V. 2022.
\newblock Resolution-robust large mask inpainting with fourier convolutions.
\newblock In \emph{Proceedings of the IEEE/CVF winter conference on applications of computer vision}, 2149--2159.

\bibitem[{Tang et~al.(2023)Tang, Zhang, Chen, Wang, and Furukawa}]{mvdiffusion}
Tang, S.; Zhang, F.; Chen, J.; Wang, P.; and Furukawa, Y. 2023.
\newblock MVDiffusion: Enabling Holistic Multi-view Image Generation with Correspondence-Aware Diffusion.
\newblock In \emph{Advances in Neural Information Processing Systems 36: Annual Conference on Neural Information Processing Systems 2023, NeurIPS 2023, New Orleans, LA, USA, December 10 - 16, 2023}.

\bibitem[{Wasserman et~al.(2025)Wasserman, Rotstein, Ganz, and Kimmel}]{paintbyinpaint}
Wasserman, N.; Rotstein, N.; Ganz, R.; and Kimmel, R. 2025.
\newblock Paint by Inpaint: Learning to Add Image Objects by Removing Them First.
\newblock In \emph{{IEEE/CVF} Conference on Computer Vision and Pattern Recognition, {CVPR} 2025, Nashville, TN, USA, June 11-15, 2025}, 18313--18324.

\bibitem[{Wu, Zheng, and Cham(2023)}]{lpo}
Wu, T.; Zheng, C.; and Cham, T. 2023.
\newblock {IPO-LDM:} Depth-aided 360-degree Indoor {RGB} Panorama Outpainting via Latent Diffusion Model.
\newblock \emph{arXiv preprint arXiv:2307.03177}.

\bibitem[{Wu, Zheng, and Cham(2024)}]{panodiffusion}
Wu, T.; Zheng, C.; and Cham, T. 2024.
\newblock PanoDiffusion: 360-degree Panorama Outpainting via Diffusion.
\newblock In \emph{The Twelfth International Conference on Learning Representations, {ICLR} 2024, Vienna, Austria, May 7-11, 2024}.

\bibitem[{Xiao et~al.(2024)Xiao, Wu, Xu, Dai, Hu, Lu, Zeng, Liu, and Yuan}]{florence2}
Xiao, B.; Wu, H.; Xu, W.; Dai, X.; Hu, H.; Lu, Y.; Zeng, M.; Liu, C.; and Yuan, L. 2024.
\newblock Florence-2: Advancing a Unified Representation for a Variety of Vision Tasks.
\newblock In \emph{{IEEE/CVF} Conference on Computer Vision and Pattern Recognition, {CVPR} 2024, Seattle, WA, USA, June 16-22, 2024}, 4818--4829.

\bibitem[{Xiao et~al.(2025)Xiao, Wang, Zhou, Yuan, Xing, Yan, Li, Wang, Huang, and Liu}]{omnigen}
Xiao, S.; Wang, Y.; Zhou, J.; Yuan, H.; Xing, X.; Yan, R.; Li, C.; Wang, S.; Huang, T.; and Liu, Z. 2025.
\newblock Omnigen: Unified image generation.
\newblock In \emph{Proceedings of the Computer Vision and Pattern Recognition Conference}, 13294--13304.

\bibitem[{Xu et~al.(2025)Xu, Kong, Wang, Pan, Lin, and Liu}]{insightedit}
Xu, Y.; Kong, J.; Wang, J.; Pan, X.; Lin, B.; and Liu, Q. 2025.
\newblock Insightedit: Towards better instruction following for image editing.
\newblock In \emph{Proceedings of the Computer Vision and Pattern Recognition Conference}, 2694--2703.

\bibitem[{Yang et~al.(2025{\natexlab{a}})Yang, Li, Yang, Zhang, Hui, Zheng, Yu, Gao, Huang, Lv et~al.}]{yang2025qwen3}
Yang, A.; Li, A.; Yang, B.; Zhang, B.; Hui, B.; Zheng, B.; Yu, B.; Gao, C.; Huang, C.; Lv, C.; et~al. 2025{\natexlab{a}}.
\newblock Qwen3 technical report.
\newblock \emph{arXiv preprint arXiv:2505.09388}.

\bibitem[{Yang et~al.(2025{\natexlab{b}})Yang, Duan, Zhu, Liu, Liu, Xu, Ma, Min, Zhai, and Callet}]{omni2}
Yang, L.; Duan, H.; Zhu, Y.; Liu, X.; Liu, L.; Xu, Z.; Ma, G.; Min, X.; Zhai, G.; and Callet, P.~L. 2025{\natexlab{b}}.
\newblock Omni$^2$: Unifying Omnidirectional Image Generation and Editing in an Omni Model.
\newblock \emph{arXiv preprint arXiv:2504.11379}.

\bibitem[{Zhang et~al.(2024)Zhang, Wu, Gambardella, Huang, Phung, Ouyang, and Cai}]{panfusion}
Zhang, C.; Wu, Q.; Gambardella, C.~C.; Huang, X.; Phung, D.; Ouyang, W.; and Cai, J. 2024.
\newblock Taming stable diffusion for text to 360 $\{$$\backslash$deg$\}$ panorama image generation.
\newblock \emph{arXiv preprint arXiv:2404.07949}.

\bibitem[{Zhang et~al.(2023)Zhang, Mo, Chen, Sun, and Su}]{magicbrush}
Zhang, K.; Mo, L.; Chen, W.; Sun, H.; and Su, Y. 2023.
\newblock MagicBrush: {A} Manually Annotated Dataset for Instruction-Guided Image Editing.
\newblock In \emph{Advances in Neural Information Processing Systems 36: Annual Conference on Neural Information Processing Systems 2023, NeurIPS 2023, New Orleans, LA, USA, December 10 - 16, 2023}.

\bibitem[{Zhang et~al.(2018)Zhang, Isola, Efros, Shechtman, and Wang}]{lpips}
Zhang, R.; Isola, P.; Efros, A.~A.; Shechtman, E.; and Wang, O. 2018.
\newblock The unreasonable effectiveness of deep features as a perceptual metric.
\newblock In \emph{Proceedings of the IEEE conference on computer vision and pattern recognition}, 586--595.

\bibitem[{Zhang et~al.(2014)Zhang, Song, Tan, and Xiao}]{panocontext}
Zhang, Y.; Song, S.; Tan, P.; and Xiao, J. 2014.
\newblock Panocontext: A whole-room 3d context model for panoramic scene understanding.
\newblock In \emph{European conference on computer vision}, 668--686.

\bibitem[{Zhang et~al.(2021)Zhang, Zhang, Lai, and Zhu}]{efficient}
Zhang, Y.; Zhang, F.; Lai, Y.; and Zhu, Z. 2021.
\newblock Efficient propagation of sparse edits on 360{\(\circ\)} panoramas.
\newblock \emph{Comput. Graph.}, 96: 61--70.

\bibitem[{Zhang et~al.(2022)Zhang, Zhang, Zhu, Wang, and Jin}]{fast}
Zhang, Y.; Zhang, F.; Zhu, Z.; Wang, L.; and Jin, Y. 2022.
\newblock Fast Edit Propagation for 360 Degree Panoramas Using Function Interpolation.
\newblock \emph{{IEEE} Access}, 10: 43882--43894.

\bibitem[{Zhang et~al.(2025)Zhang, Xie, Lu, Yang, and Yang}]{icedit}
Zhang, Z.; Xie, J.; Lu, Y.; Yang, Z.; and Yang, Y. 2025.
\newblock In-Context Edit: Enabling Instructional Image Editing with In-Context Generation in Large Scale Diffusion Transformer.
\newblock \emph{arXiv preprint arXiv:2504.20690}.

\bibitem[{Zhao et~al.(2024)Zhao, Ma, Chen, Si, Wu, An, Yu, Zhang, Li, and Chang}]{ultraedit}
Zhao, H.; Ma, X.~S.; Chen, L.; Si, S.; Wu, R.; An, K.; Yu, P.; Zhang, M.; Li, Q.; and Chang, B. 2024.
\newblock Ultraedit: Instruction-based fine-grained image editing at scale.
\newblock \emph{Advances in Neural Information Processing Systems}, 37: 3058--3093.

\bibitem[{Zheng et~al.(2020)Zheng, Zhang, Li, Tang, Gao, and Zhou}]{structured3d}
Zheng, J.; Zhang, J.; Li, J.; Tang, R.; Gao, S.; and Zhou, Z. 2020.
\newblock Structured3d: A large photo-realistic dataset for structured 3d modeling.
\newblock In \emph{European Conference on Computer Vision}, 519--535. Springer.

\end{thebibliography}

\clearpage 

\twocolumn[{%
	\centering
	
	\Huge \textbf{Supplementary Material} \\[15pt] 
	
	\LARGE SE360: Semantic Edit in 360° Panoramas via Hierarchical Data Construction \\[30pt] 
}]

\section{Additional Experimental Results}
To more comprehensively demonstrate the effectiveness of our method, this section provides additional qualitative and quantitative experimental results.

\subsection{Additional Qualitative Comparisons}

\begin{figure*}[t!]
\centering
\includegraphics[width=2.00\columnwidth]{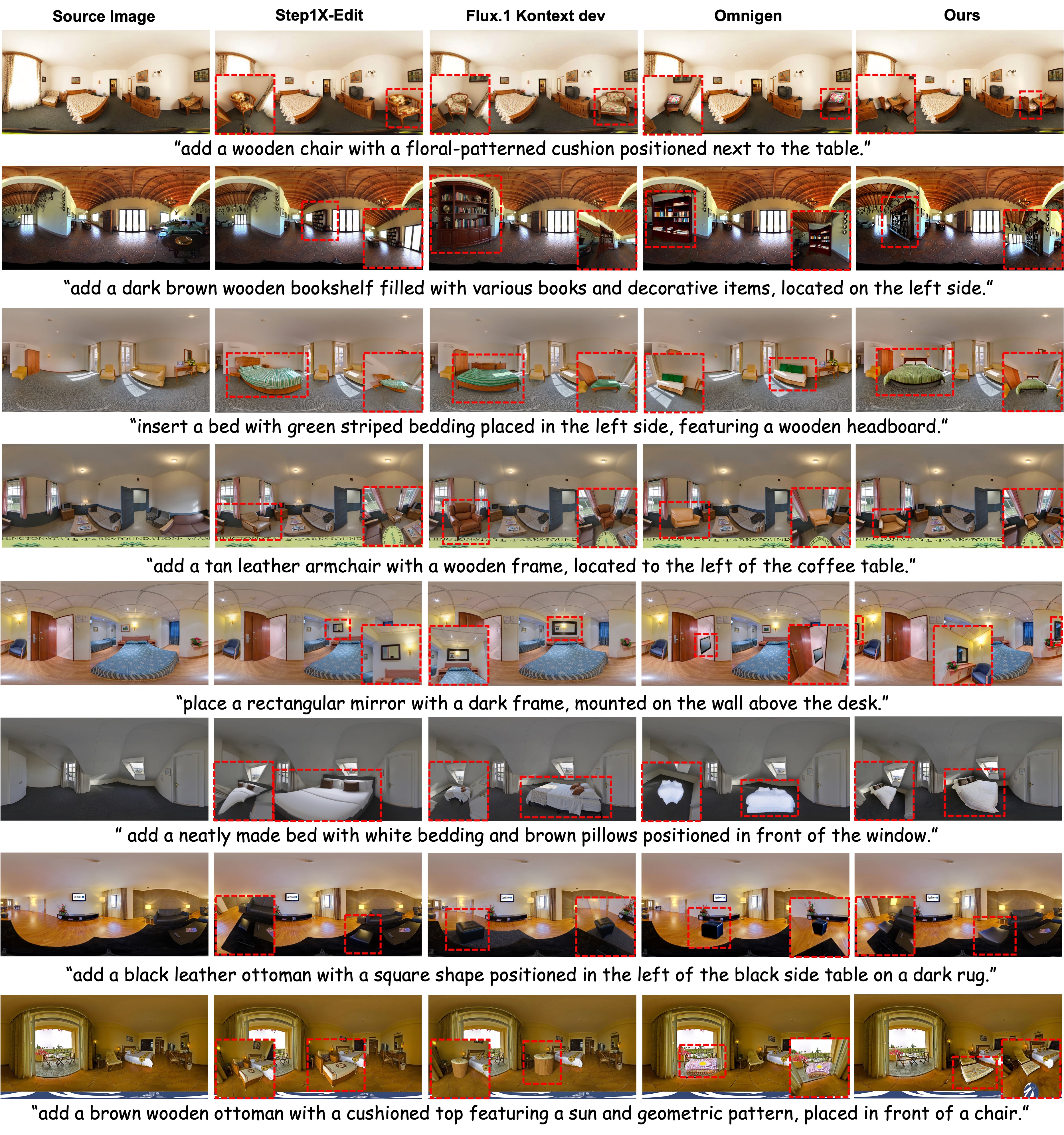} 
\caption{The comparation with existing models in addition task.}
\label{figs1}
\end{figure*}

As illustrated in Figure~\ref{figs1}, we present a further qualitative comparison of the outputs from our method against existing approaches on the object insertion task. The editing results reveal that existing methods suffer from significant degradation when processing 360° panoramic inputs. Objects generated in the Equirectangular Projection (ERP) often lack geometric consistency when projected into perspective views. Moreover, these methods frequently fail to comprehend object semantics, especially near the image boundaries, leading to an inability to accurately adhere to the editing instructions. In contrast, our method not only exhibits excellent consistency in perspective views and across the left-right boundaries, but the generated objects are also aware of the ambient illumination, rendering plausible reflections, shading, and cast shadows.

\begin{figure*}[t!]
\centering
\includegraphics[width=2.00\columnwidth]{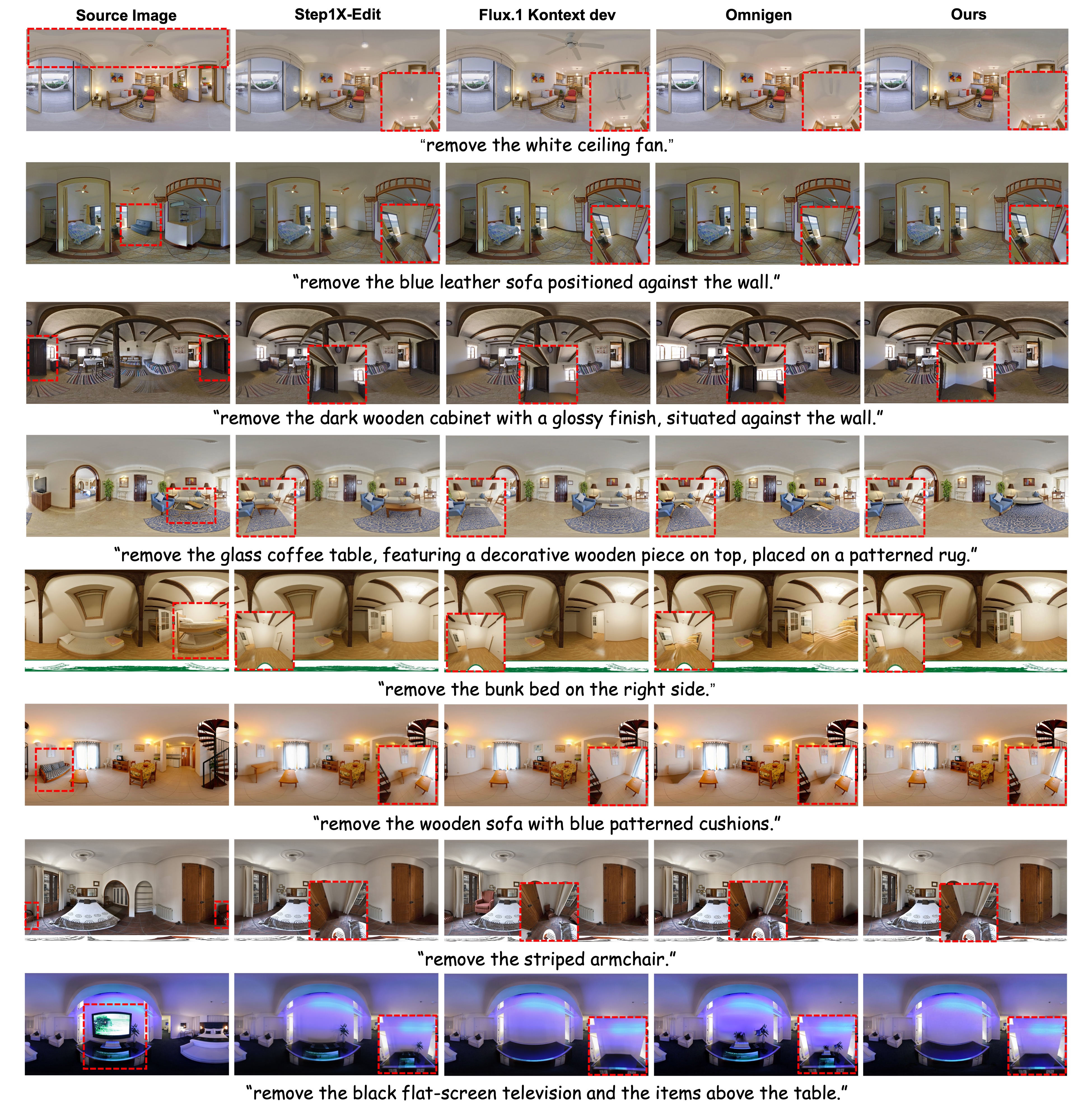} 
\caption{The comparation with existing models in removal task.}
\label{figs2}
\end{figure*}

Furthermore, existing methods exhibit similar performance degradation in object removal tasks, particularly at the image boundaries, as depicted in Figure~\ref{figs2}. Even when simultaneously removing objects straddling the left and right edges, the subsequent inpainting often introduces noticeable seams and artifacts, disrupting the continuity of the panoramic scene. Conversely, our method not only overcomes this boundary continuity challenge but also adeptly removes the object and its corresponding shadow in conjunction.

\subsection{Detailed Quantitative Results}
To provide a more in-depth analysis, the tables below present a detailed breakdown of the averaged results reported in the main paper across the two test sets: PanoEval and PanoEval-Boundary.

Table~\ref{tab:appendix_addition} and Table~\ref{tab:appendix_removal} show the detailed performance metrics of our model against the baselines on the object addition and removal tasks, respectively. The data indicates that our SE360 model achieves state-of-the-art or competitive performance across most key metrics in both standard scenarios (PanoEval) and more challenging boundary scenarios (PanoEval-Boundary).

\begin{table*}[t]
\centering
\small
\setlength{\tabcolsep}{5pt}
\begin{tabular}{l ccc ccc}
\toprule
& \multicolumn{3}{c}{\textbf{PanoEval}} & \multicolumn{3}{c}{\textbf{PanoEval-Boundary}} \\
\cmidrule(lr){2-4} \cmidrule(lr){5-7}
\textbf{Model} & CS$_{\text{pers}}$$\uparrow$ & LPIPS$\downarrow$ & LPIPS$_{\text{pers}}$$\downarrow$ & CS$_{\text{pers}}$$\uparrow$ & LPIPS$\downarrow$ & LPIPS$_{\text{pers}}$$\downarrow$ \\
\midrule
InstructPix2Pix & 23.520 & 0.783 & 0.797 & 24.075 & 0.782 & 0.798 \\
Erasedraw & 25.259 & 0.732 & 0.736 & 25.271 & 0.737 & 0.740 \\
ICEdit & 27.282 & 0.283 & 0.354 & 27.002 & 0.294 & 0.360 \\
MagicBrush & 27.621 & 0.107 & 0.214 & 27.610 & 0.113 & 0.232 \\
PaintByInpaint & 28.082 & 0.093 & 0.194 & 28.217 & 0.095 & 0.206 \\
OmniGen & 28.488 & 0.088 & 0.203 & 28.106 & 0.097 & 0.224 \\
Flux.1 Kontext dev & \textbf{29.666} & 0.137 & 0.264 & 28.464 & 0.149 & 0.268 \\
Step1X-Edit & 29.254 & 0.087 & 0.209 & 28.170 & 0.101 & 0.213 \\
\midrule
\textbf{SE360 (Ours)} & 29.307 & \textbf{0.075} & \textbf{0.173} & \textbf{29.459} & \textbf{0.077} & \textbf{0.185} \\
\bottomrule
\end{tabular}
\caption{Quantitative comparison for the object addition task on the two test sets.}
\label{tab:appendix_addition}
\end{table*}

\begin{table*}[t]
\centering
\small 
\setlength{\tabcolsep}{3.5pt} 
\begin{tabular}{l ccccc ccccc}
\toprule
& \multicolumn{5}{c}{\textbf{PanoEval}} & \multicolumn{5}{c}{\textbf{PanoEval-Boundary}} \\
\cmidrule(lr){2-6} \cmidrule(lr){7-11}
\textbf{Model} & CS-No$_{\text{pers}}$$\uparrow$ & FAED$\downarrow$ & LPIPS$\downarrow$ & LPIPS$_{\text{pers}}$$\downarrow$ & PSNR$\uparrow$ & CS-No$_{\text{pers}}$$\uparrow$ & FAED$\downarrow$ & LPIPS$\downarrow$ & LPIPS$_{\text{pers}}$$\downarrow$ & PSNR$\uparrow$ \\
\midrule
MagicBrush & 71.346 & 2.256 & 0.116 & 0.216 & 20.381 & 71.258 & 4.013 & 0.127 & 0.237 & 20.016 \\
ICEdit & 72.248 & 1.298 & 0.265 & 0.307 & 23.496 & 72.197 & 2.101 & 0.276 & 0.320 & 22.077 \\
OmniGen & 72.517 & 0.784 & {0.062} & {0.130} & 25.546 & 71.669 & 1.020 & {0.075} & {0.168} & 24.423 \\
Flux.1 Kontext dev & 72.020 & 1.371 & 0.106 & 0.176 & 20.190 & 71.876 & 2.049 & 0.134 & 0.225 & 19.217 \\
Step1X-Edit & {72.659} & {0.661} & \textbf{0.053} & \textbf{0.086} & \textbf{28.979} & {72.488} & {0.943} & \textbf{0.064} & \textbf{0.113} & \textbf{27.332} \\
\midrule
\textbf{SE360 (Ours)} & \textbf{73.126} & \textbf{0.271} & 0.076 & 0.175 & {26.536} & \textbf{73.181} & \textbf{0.411} & 0.085 & 0.187 & {26.076} \\
\bottomrule
\end{tabular}
\caption{Quantitative comparison for the object removal task on the two test sets.}
\label{tab:appendix_removal}
\end{table*}

\section{More Results for the Object Addition Task}
To further demonstrate the powerful capabilities and robustness of our SE360 model on the object addition task, this section provides a wider variety of generated results. These examples cover different object categories, complex lighting conditions, and diverse background environments.

\begin{figure*}[t!]
\centering
\includegraphics[width=1.90\columnwidth]{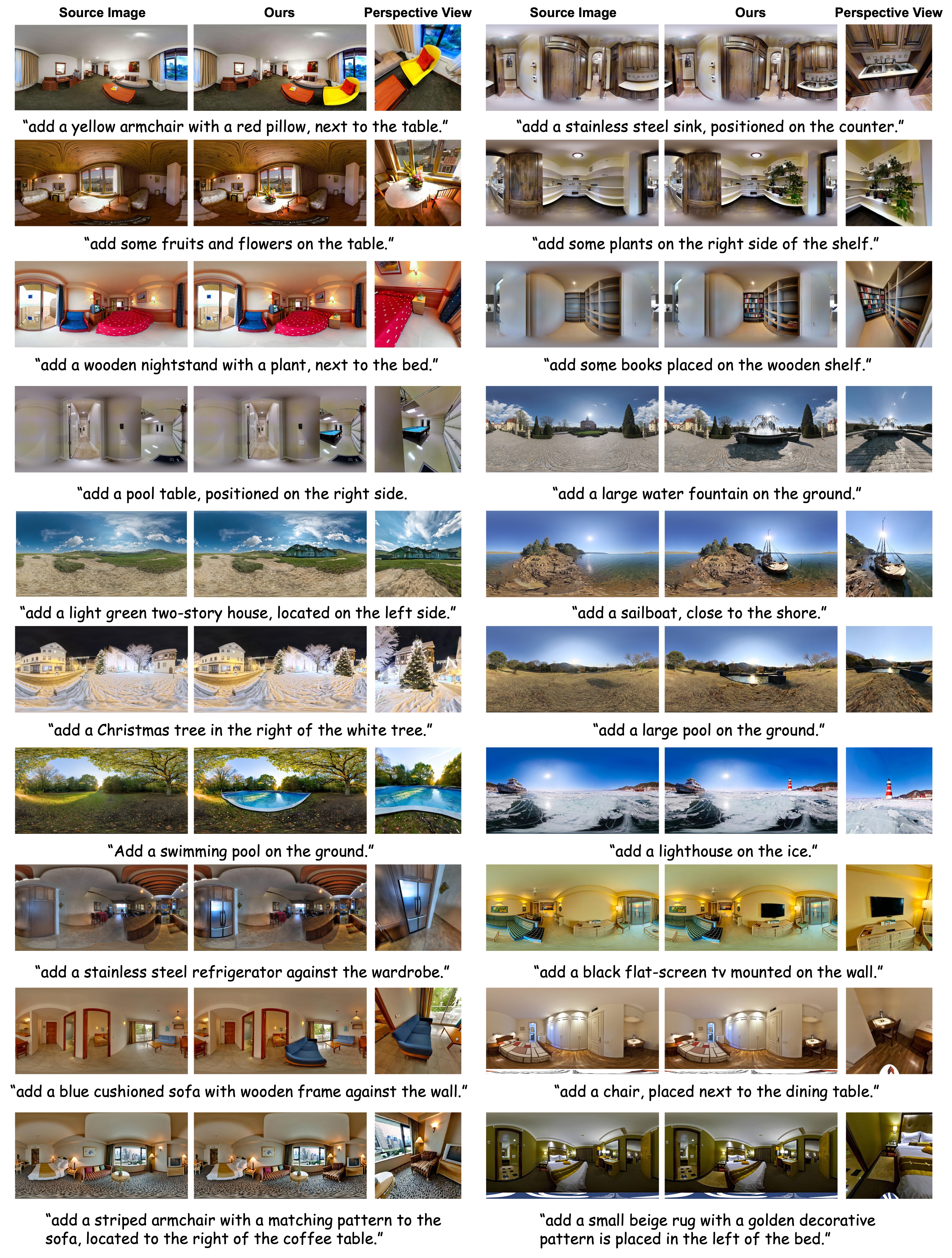} 
\caption{Additional object addition results of our model.}
\label{figs3}
\end{figure*}

As shown in Figure~\ref{figs3}, our model can accurately interpret text prompts and seamlessly integrate objects into the 360° panoramas. The generated objects are not only geometrically correct in perspective but also interact naturally with the ambient lighting, such as casting accurate reflections on glossy surfaces or rendering realistic shadows according to the environmental light source. This greatly enhances the realism and immersive quality of the scene.

\section{Results for Ablation Studies}
\begin{figure*}[t!]
\centering
\includegraphics[width=2.00\columnwidth]{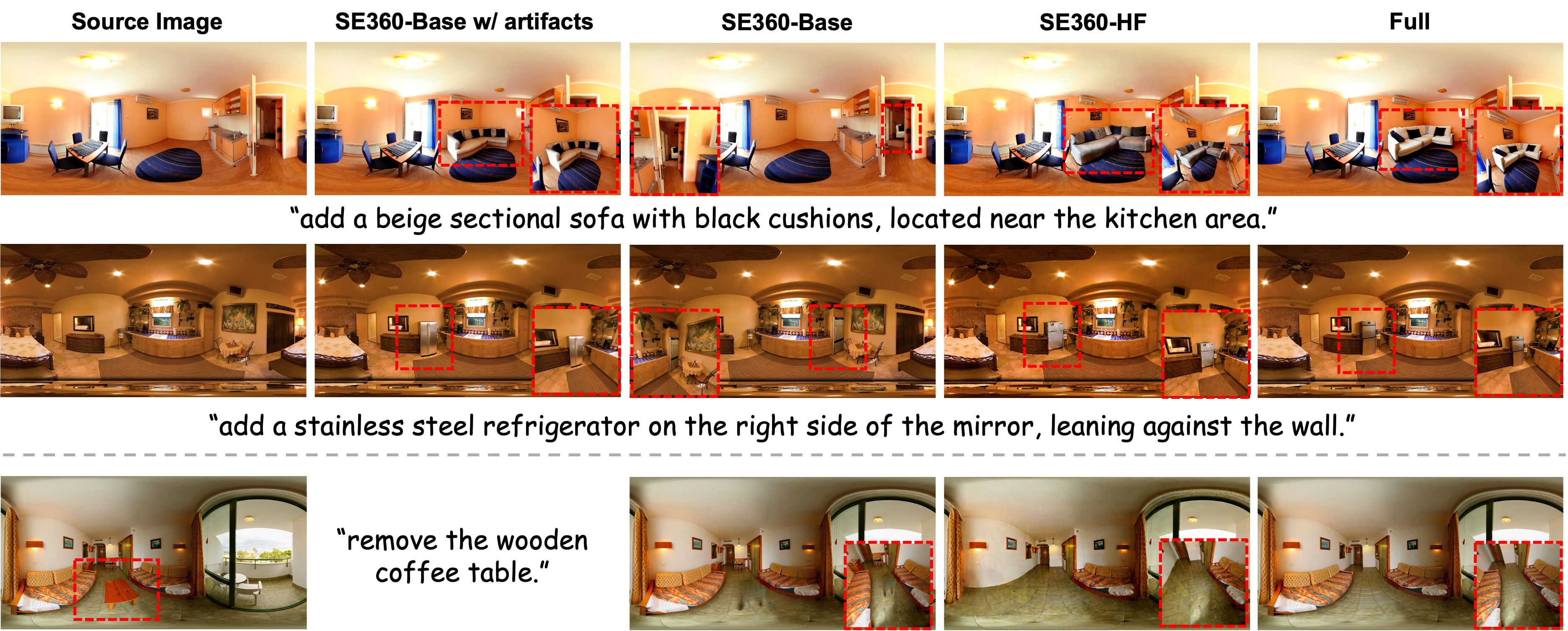} 
\caption{Ablation analysis under different training strategies.}
\label{figs4}
\end{figure*}

The ablation study in the main paper (see Table 2 in main paper) reveals the critical impact of different training datasets on model performance. This section provides a more in-depth explanation and presents qualitative results in Figure~\ref{figs4} to substantiate our conclusions.

We observed that mask-based erasing methods are sensitive to the context surrounding the mask. When an object is in a complex environment, such as one involving its own shadow or occlusions from other objects, these methods often produce artifacts. Since our data sources, Matterport3D and Structured3D, contain numerous scenes with complex lighting and occlusions, the SE360-Base dataset inevitably includes some data with such erasing artifacts.

As illustrated in the example of SE360-Base in Figure~\ref{figs8}, this data characteristic significantly influences the model's behavior:

\begin{itemize}
    \item \textbf{In the Object Addition Task}:
    \begin{itemize}
        \item \textbf{Model trained only on SE360-Base}: When the input image contains erasing artifacts from LaMa (mimicking the training data distribution), this model performs well in generating an object at the artifact's location. However, when the input is a clean image without artifacts, the object placement becomes more random, indicating that the model has overfit to the artifacts in the training data.
        \item \textbf{Model trained only on SE360-HF}: As SE360-HF is a small, high-quality dataset, the model shows better consistency in placement but lacks fine details and realism in the generated objects due to insufficient training data.
    \end{itemize}
    
    \item \textbf{In the Object Removal Task}:
    \begin{itemize}
        \item \textbf{Model trained only on SE360-Base}: After removing an object, this model tends to generate artifacts in its place, a behavior consistent with the output of the LaMa erasing model used to create its training data.
        \item \textbf{Model trained only on SE360-HF}: Due to the limited training data, this model sometimes exhibits "over-erasing," removing nearby objects not specified in the prompt. While this may unintentionally clear more irrelevant elements, it also explains its high CS-No score, as it might produce a more "empty" scene.
    \end{itemize}
\end{itemize}

In summary, our two-stage training strategy—first pre-training on the large-scale SE360-Base and then fine-tuning on the high-quality SE360-HF—is crucial for balancing the model's generative capabilities, robustness, and high-fidelity output. Considering the significant cost of using advanced editing models like Flux.1 Kontext max (\$0.08 per image), we strategically apply it only to a small subset of data in our refinement stage. This approach represents a highly cost-effective solution, achieving high-quality results without incurring prohibitive expenses.

\section{Multi-Condition Guided Editing Capabilities}
\begin{figure}[t!]
\centering
\includegraphics[width=1.00\columnwidth]{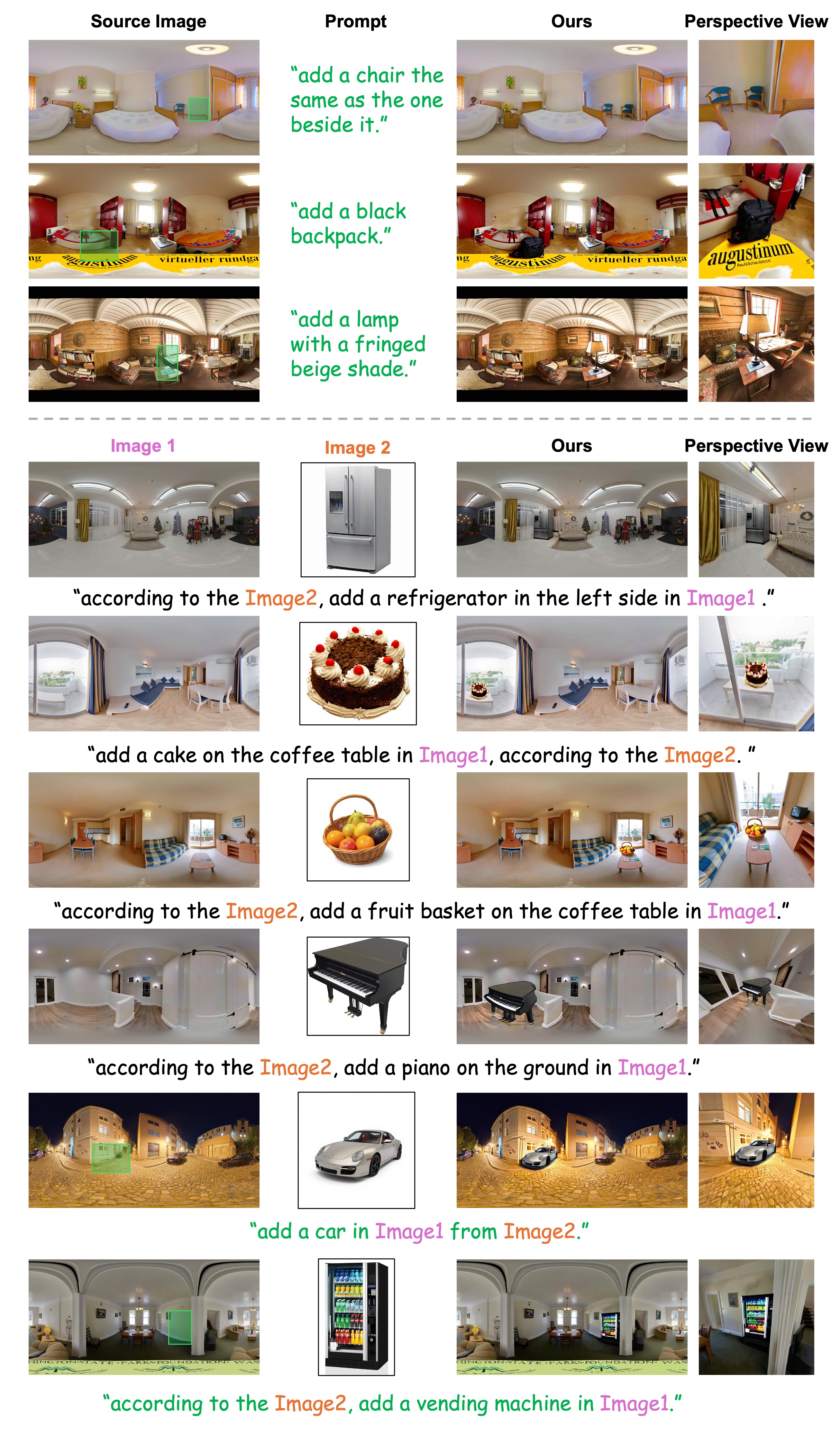} 
\caption{Output demonstration under multiple conditional input. Green means edit with mask map.}
\label{figs5}
\end{figure}

In addition to text-prompt-based editing, SE360 can be guided by multiple conditions to enable more precise user control. As shown in Figure~\ref{figs5}, a user can provide a mask map (green box) to specify the exact region for an edit, or input a reference image to constrain the appearance, or texture of the generated object.

Our model effectively fuses these inputs from different modalities to produce high-quality results that adhere to the textual description while strictly following spatial and appearance constraints. This flexibility in multi-condition input significantly expands the creative potential and user control of our framework in practical applications.

\section{Multi-Turn Editing Capability}
In many real-world scenarios, a user may need to make several sequential modifications to a scene. This requires a model that can not only edit an original image but also robustly edit its own generated output without quality degradation or artifact accumulation.

\begin{figure}[t!]
\centering
\includegraphics[width=1.00\columnwidth]{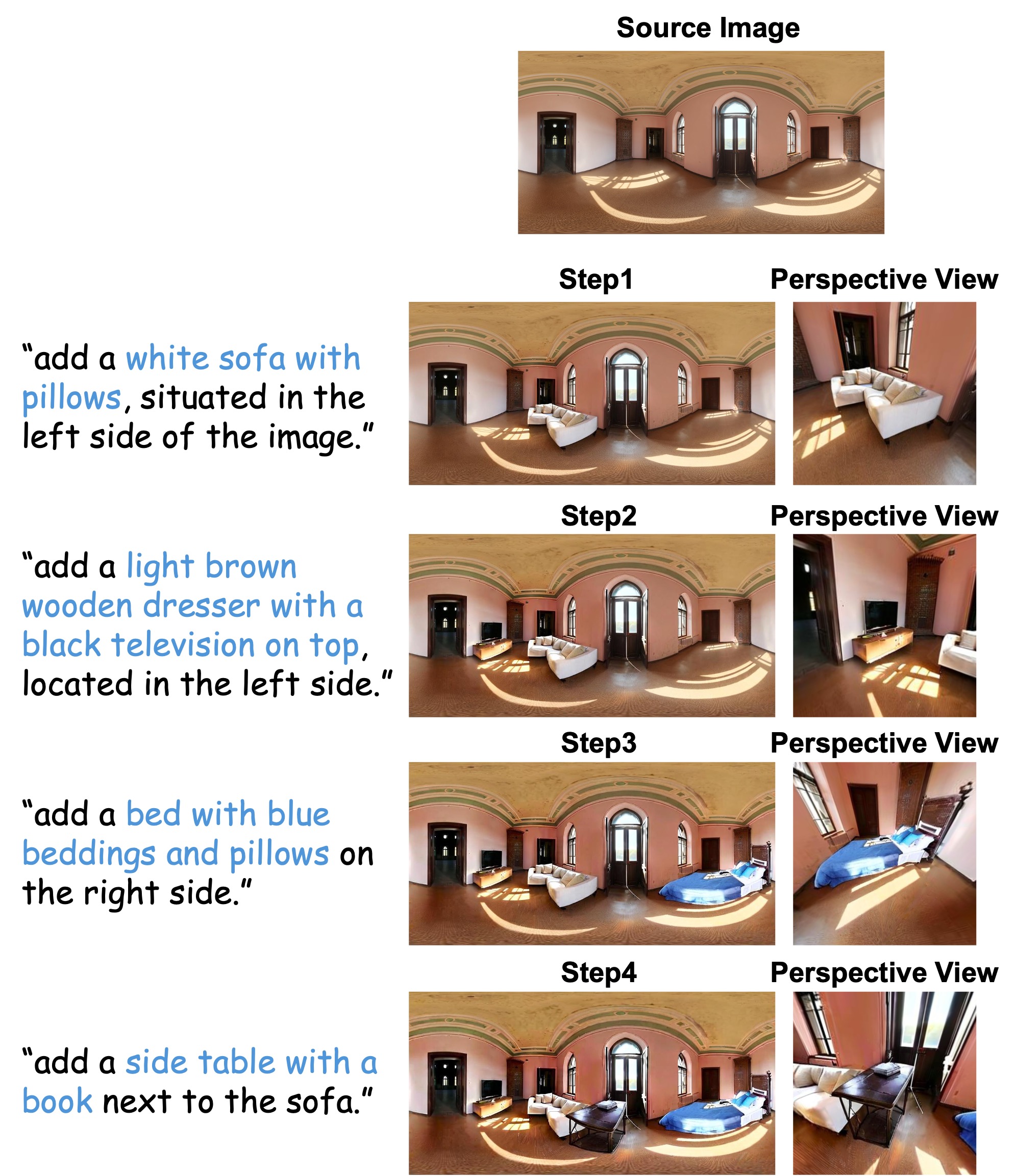} 
\caption{Multi-turn edit results of SE360.}
\label{figs6}
\end{figure}

We demonstrate the multi-turn editing capability of SE360 in Figure~\ref{figs6}. Starting with an original panorama, we first perform an "object addition" operation. We then use the resulting image as the input for a second edit. The results show that SE360 maintains a high degree of image consistency and realism even after sequential edits, proving its stability and utility in complex, progressive creative tasks.

\section{More Details of the Data Generation Pipeline}

Our data generation process is at the core of the SE360 framework and consists of two main phases: (1) \textbf{SE360-Base}, for generating a large-scale base dataset, and (2) \textbf{SE360-HF}, for high-fidelity refinement of a subset of the data. This section elaborates on these phases, particularly the five core stages of SE360-Base. The example of SE360-Base and SE360-HF are shown in Figure~\ref{figs8}.

\subsection{SE360-Base: Large-Scale Base Dataset Generation}

The goal of the SE360-Base phase is to automatically construct large-scale image editing pairs from unlabeled panoramic images. This process strictly follows the five stages outlined below:

\subsubsection{Stage 1: Object Extraction}
In this stage, we use Qwen2.5-VL-32B as VLM. Through carefully engineered prompts, we guide the model to identify and describe the primary foreground objects within each scene. This initial step yields a structured list of objects, each annotated with a detailed description and a concise category, establishing the semantic groundwork for subsequent localization tasks. The obejct description prompt in ERP, shown in Listing~\ref{lst:vlm_prompt_objcaption_generation}, explicitly instructs the model to return the output in a JSON format for easy parsing.

Subsequently, we adopted a multi-model fusion strategy to achieve robust phrase grounding, leveraging the complementary strengths of three distinct models: an initial VLM, Grounding DINO, and Florence-2. The vlm grounding prompt in ERP, shown in Listing~\ref{lst:vlm_prompt_objgrounding}.

The core of our fusion logic is a consensus-based filtering mechanism. For each object description generated by the VLM, we concurrently employ both Grounding DINO (BBox threshold=0.3, text threshold=0.25) and Florence-2 for localization. A detection is deemed valid only if a candidate bounding box from either Grounding DINO or Florence-2 exhibits an Intersection over Union (IoU) greater than 0.5 with the VLM's initial anchor detection. Among all valid candidates for the same object, we select the one with the largest bounding box area to ensure the most comprehensive coverage. This process effectively ensures that each accepted localization is endorsed by a "vote" from at least two models.

For objects that the VLM initially failed to localize but were detected by both Grounding DINO and Florence-2, we apply a similar consensus strategy, retaining high-confidence DINO boxes only if they have a significant overlap ($IoU > 0.5$) with their corresponding Florence-2 detections.

Finally, to refine the output, we apply Non-Maximum Suppression (NMS) with an IoU threshold of 0.3 and a series of geometric filters to eliminate redundant detections and spurious bounding boxes, thereby ensuring the precision and fidelity of the final annotations. These geometric filters, applied to the Florence-2 detections prior to fusion, discard bounding boxes that meet any of the following criteria: 1) touching the image boundary (within a 30-pixel margin), 2) having a longest side that exceeds 60\% of the image height, 3) covering less than 0.3\% of the total image area, or 4) covering more than 40\% of the total image area.

\subsubsection{Stage 2: Physical Affiliation Analysis}
A key innovation within the SE360-Base pipeline is the Physical Affiliation Analysis (PAA) module. We found that directly using BBox-guided segmentation methods like Grounded-SAM often leads to incomplete segmentation of composite objects. To address this, the PAA module is designed to identify and include these physically affiliated items. It first uses a Vision-Language Model (VLM) to identify the core object, then explicitly queries for all associated sub-items through targeted prompts (e.g., "what is on/inside/attached to the item?"). In this way, we can gather a complete object group and generate a comprehensive and semantically whole segmentation mask for it.

\begin{figure}[t!]
\centering
\includegraphics[width=1.00\columnwidth]{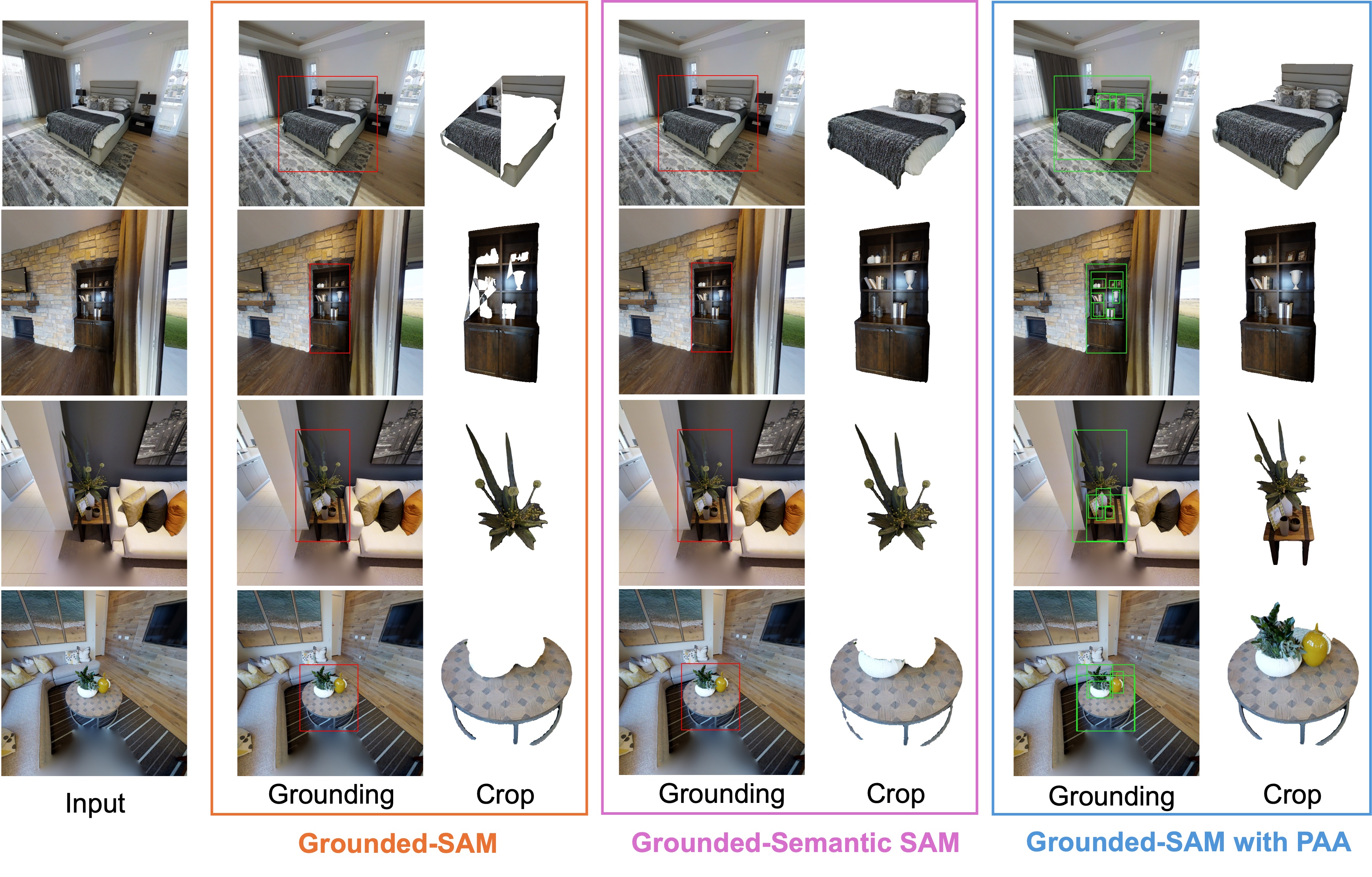} 
\caption{Extra comparison with different segmentation methods. Grounding provided by Grounding-Dino, Crop provided by SAM or Semantic SAM.}
\label{figs7}
\end{figure}

\begin{figure*}[t!]
\centering
\includegraphics[width=2.00\columnwidth]{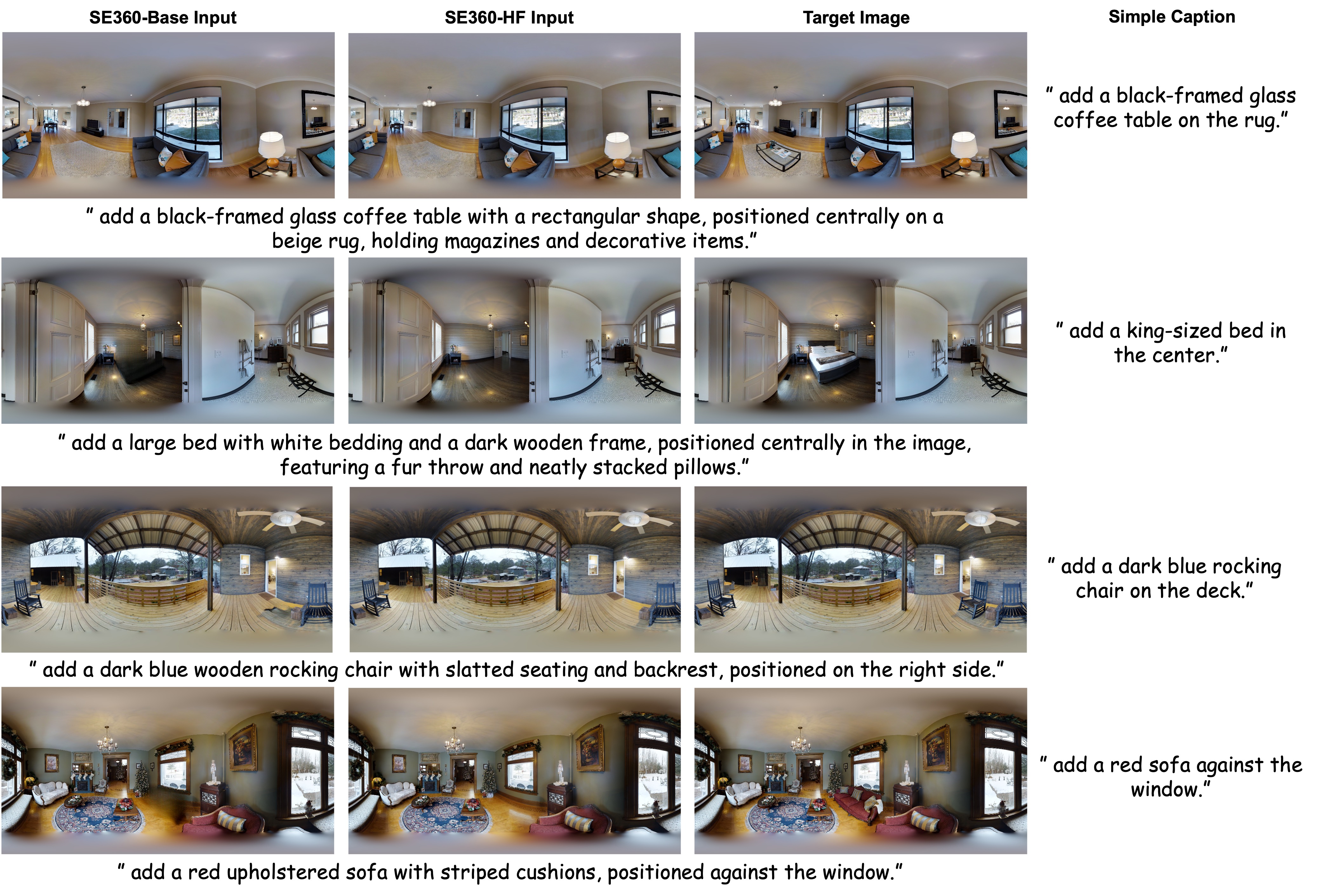}
\caption{The demonstration of SE360-Base and SE360-HF dataset in addition task.}
\label{figs8}
\end{figure*}

Figure~\ref{figs7} visually demonstrates the effectiveness of the PAA module. Without PAA, the segmentation results are incomplete, even when using models like SAM or Semantic-SAM. When PAA is enabled, however, our system generates a holistic mask that covers both the core object and all its affiliated items. This is crucial for creating high-quality training pairs and preventing illogical editing outcomes. The VLM prompt for PAA, shown in Listing~\ref{lst:vlm_prompt_PAA}.

\subsubsection{Stage 3: Adaptive Projection Adjustment}
Following the acquisition of an accessory item list, we perform fine-grained detection within an adaptively adjusted perspective viewport. Our approach diverges from static projection by dynamically optimizing the FOV and projection center to ensure all target objects are rendered with complete and clear visibility.

To better focus on target objects, the framework applies a vertical offset to the projection center, effectively "elevating" the virtual camera's viewpoint. This offset is calculated based on the relative height of each object's bounding box. The offset coefficient is inversely proportional to the object's height, ranging from 0.05 to 0.4 times the bounding box height, which grants smaller objects a more pronounced perspective elevation.

Initially, we project an optimized perspective view and employ Grounding DINO (box threshold: 0.25, text threshold: 0.25) to detect primary and accessory items. If any detected bounding box intersects the viewport boundary (within a 30-pixel margin)—indicating partial object visibility—the framework initiates an iterative refinement process. It multiplies the FOV expansion factor by 1.15 and further increases the vertical offset factor by 1.1 to adjust the projection center. This loop persists until all detected objects are fully contained within the viewport or a predefined maximum of 3 iterations is reached. This procedure guarantees that subsequent segmentation operations are performed on complete object views. Such an adaptive mechanism is crucial for managing the complex spatial arrangements inherent in 360° panoramic scenes.

\subsubsection{Stage 4: Mask-Guided Erasing}
For mask-guided erasure, we employ a progressive segmentation strategy based on SAM2. Initially, a coarse mask is generated from the object's bounding box. This mask is subsequently refined by re-prompting SAM2 with five positive points sampled from its interior. These five points consist of the mask's geometric centroid, as well as the leftmost, rightmost, topmost, and bottommost points on the axes passing through it.

Specifically, the algorithm first computes the bounding box of the mask region to identify its geometric centroid. If the centroid falls outside the mask, the nearest pixel within the mask is selected as a substitute. Four boundary points are then extracted along the horizontal and vertical lines that intersect at this central point, forming a cruciform sampling pattern.

Subsequently, we extract the centroids from all detected masks and utilize them as point prompts for a third prediction round to further optimize and merge masks. The resulting high-precision mask undergoes post-processing, where noisy regions with an area smaller than 30 pixels are removed, and the mask is dilated with a $15 \times 15$ pixel kernel to create an erasure buffer.

Critically, instead of performing erasure on isolated perspective image patches, we reproject this final mask back into the complete ERP coordinate space. This ERP-space mask then guides the LaMa inpainter to perform erasure directly on the original panorama. This approach fully leverages global contextual information, thereby achieving more coherent and high-fidelity inpainting results.

\subsubsection{Stage 5: Instruction Recaption}
To enhance the descriptive richness of our dataset in support of downstream instruction-editing tasks, we introduce an Instruction Recaption module. Leveraging the "original description" from Stage 1 and the clear, adaptively-adjusted perspective view from the preceding step, this module utilizes a VLM to generate two new, hierarchical descriptions:
\begin{enumerate}
\item A Standard Refined Description, which integrates verifiable visual details from the high-quality perspective view with a pre-established spatial context.
\item A Brief Description, which offers a succinct summary of the object's core attributes and location.
\end{enumerate}
This dual-description methodology furnishes each object instance with labels that are both detailed and context-aware, as well as concise and easily parsable, enabling models to adapt to input instructions of varying complexity. The VLM prompt for instruction recaption, shown in Listing~\ref{lst:vlm_prompt_instruction recaption}. 
Subsequently, the generated JSON file is input to an LLM, Qwen3-8B, which classifies the localization method as either absolute (relative to the image frame) or relative (with respect to other objects). This classification informs our image rotation strategy during model training. The LLM prompt for caption classification, shown in Listing~\ref{lst:llm_prompt_caption classification}.

\subsection{SE360-HF: High-Fidelity Refinement}
Object erasure in complex scenes often produces visual artifacts, such as residual shadows. This can cause the model to overfit to these defects when generating new content. Although downsampling helps mitigate such artifacts during training, this approach is less effective for large-scale objects.

To address this issue, we employ a high-fidelity data generation stage, SE360-HF. We replace LaMa with a more powerful instruction-driven editor, Flux.1 Kontext max, to produce cleaner image-edit pairs. The resulting images undergo a two-stage filtering process:

\begin{enumerate}
\item SSIM Filtering: We first compute the SSIM between the erased image and the original. Images with a score exceeding a predefined threshold of 0.86 are discarded.
\item Region-Constrained Feature Similarity Check: We perform a local similarity check using DINOv2 features within the mask's bounding box and at the image boundaries. Specifically, we compute the feature similarity within the masked region and within a 10\%-width border region of the image, respectively. Unlike a global comparison, this region-constrained strategy allows for consistent filtering of objects of varying sizes with a uniform threshold. Images are excluded if the similarity within the masked region exceeds 0.9 (indicating an insignificant edit) or if the similarity in the border region falls below 0.8 (indicating a disruption of boundary continuity).
\end{enumerate}

\section{Training Details}
Our model, SE360, contains 3.9 billion parameters and is initialized from the weights of OmniGen. The training is based on the Flow Matching framework and employs efficient fine-tuning via LoRA, with both rank and alpha set to 16. We adopt a two-stage training strategy designed to first learn general editing capabilities and then refine for high-quality output. The entire training process was conducted on 4 NVIDIA A100 GPUs using BF16 mixed-precision.

\subsection{Main Task Training (Object Addition/Removal)}
For the core tasks of object addition and removal, we train a single LoRA weight to enhance the model's versatility.

\subsubsection{Training Procedure and Hyperparameters}
The training is divided into two stages:
\begin{itemize}
    \item \textbf{Stage 1 (Foundational Training)}: We conduct 10,000 iterations on the large-scale SE360-Base dataset, using a batch size of 64 and an initial learning rate of $1 \times 10^{-3}$.
    \item \textbf{Stage 2 (Fine-tuning)}: Using the weights from stage one, we fine-tune for 1,000 iterations on the high-quality SE360-HF dataset, with a batch size of 256 and an initial learning rate of $3 \times 10^{-4}$.
\end{itemize}
For both stages, the learning rate is decayed at each step using a \textbf{Cosine Annealing} scheduler, with a minimum learning rate of $1 \times 10^{-7}$. The panoramic image resolution used during training is $1024 \times 512$, 1024 is width, 512 is height.

\subsubsection{Conditional Guidance and Dropout}
To improve the model's generalization and robustness to inputs, we use the following conditional guidance strategies:
\begin{itemize}
    \item \textbf{Mask Guidance}: To enable precise local editing, a mask map is applied with a 20\% probability.
    \item \textbf{Text Prompts}: We use detailed text prompts with a 70\% probability and brief text prompts with a 30\% probability.
    \item \textbf{Conditional Dropout}: During training, the image condition dropout rate is 1\%, and the text condition dropout rate is 1\%.
\end{itemize}

\subsubsection{Location-Aware Encoding and Decoding}
 In "Pad-and-Unpad" strategy, the padding at the pixel level is 64 pixels, while the padding in the latent space is 8 units.
    
\subsubsection{Orientation-Decoupled Data Augmentation}
To enhance the model's generalization while maintaining the accuracy of directional prompts (e.g., "in the center of the image"), we introduce an orientation-decoupled data augmentation strategy. We first classify the text prompts from the SE360-Base dataset:
        \begin{itemize}
            \item For prompts containing absolute position descriptions, the image is randomly rotated latitudinally within a small range (0 to 10 degrees).
            \item For prompts containing relative position descriptions, the image can be randomly rotated latitudinally by any angle.
        \end{itemize}

Standard data augmentations such as scaling and rotation are also applied.

\subsection{Reference-Based Addition Training}
For the reference-based object addition task, we train a separate, dedicated LoRA weight to better learn the ability to extract features from a reference image and integrate them into the scene.
\begin{itemize}
    \item \textbf{Training Setup}: The resolution of reference images is standardized to $256 \times 256$. During training, the reference image is used as a condition with a 90\% probability.
    \item \textbf{Reference Image Augmentation}: To enhance the model's adaptability to variations in the reference image, we apply a rich set of data augmentations with the following probabilities:
        \begin{itemize}
            \item Rotation Probability: 50\%
            \item Affine Transform Probability: 20\%
            \item Scaling Probability: 50\%
            \item Color Jitter Probability: 20\%
            \item Noise Injection Probability: 10\%
            \item Blurring Probability: 10\%
            \item Sharpening Probability: 10\%
            \item Gamma Correction Probability: 10\%
            \item Random Block Crop Probability: 30\%
            \item Keep Background Probability: 30\%
        \end{itemize}
\end{itemize}

\section{User Study Details}
This section provides further details on our user study, including the design of the evaluation interface and a more in-depth statistical analysis of the rating results.

\subsection{Evaluation Interface}
To allow participants to comprehensively and interactively evaluate the panoramic editing results, we designed and developed a custom web-based evaluation interface, shown in Figure~\ref{figs11}. The interface includes several key components:
\begin{itemize}
    \item \textbf{Comparison View}: In \textbf{Evaluation of a single model}, the original input image, the model-edited output and panoramic viewer are displayed side-by-side for direct comparison. Users can click panoramic viewer and drag to freely zoom, rotate, and pan their viewpoint within the 360° space, allowing them to immersively inspect the geometry, lighting, and integration of the edited object with its surroundings.
    \item \textbf{Prompt and Rating}: Below the three views, the text prompt used for the edit is clearly displayed, along with rating buttons for three criteria: Consistency, Quality, and Plausibility. Participants were asked to independently rate each result on a 5-point Likert scale for each criterion.
\end{itemize}
This design ensures that participants' judgments are not based solely on static ERP images but are made within an interactive environment that more closely approximates a real-world experience, leading to more reliable evaluations.

\subsection{Statistical Analysis and Qualitative Examples}
To more intuitively illustrate the evaluation process of our user study and the advantages of our method, we present two representative examples in Figure~\ref{figs10}. Each example juxtaposes the visual output from our method, SE360, against four baseline models, accompanied by a bar chart showing the average user ratings for that specific case. This helps to directly link the qualitative visual perception with the quantitative scoring data. From these examples, it is clear that for tasks like adding a refrigerator or a wardrobe, other models exhibit issues such as geometric distortion, poor environmental integration, or a failure to fully adhere to the prompt. In contrast, SE360's results excel across all dimensions, and its scores lead accordingly in the case-specific charts.

\begin{figure}[t!]
\centering
\includegraphics[width=1.00\columnwidth]{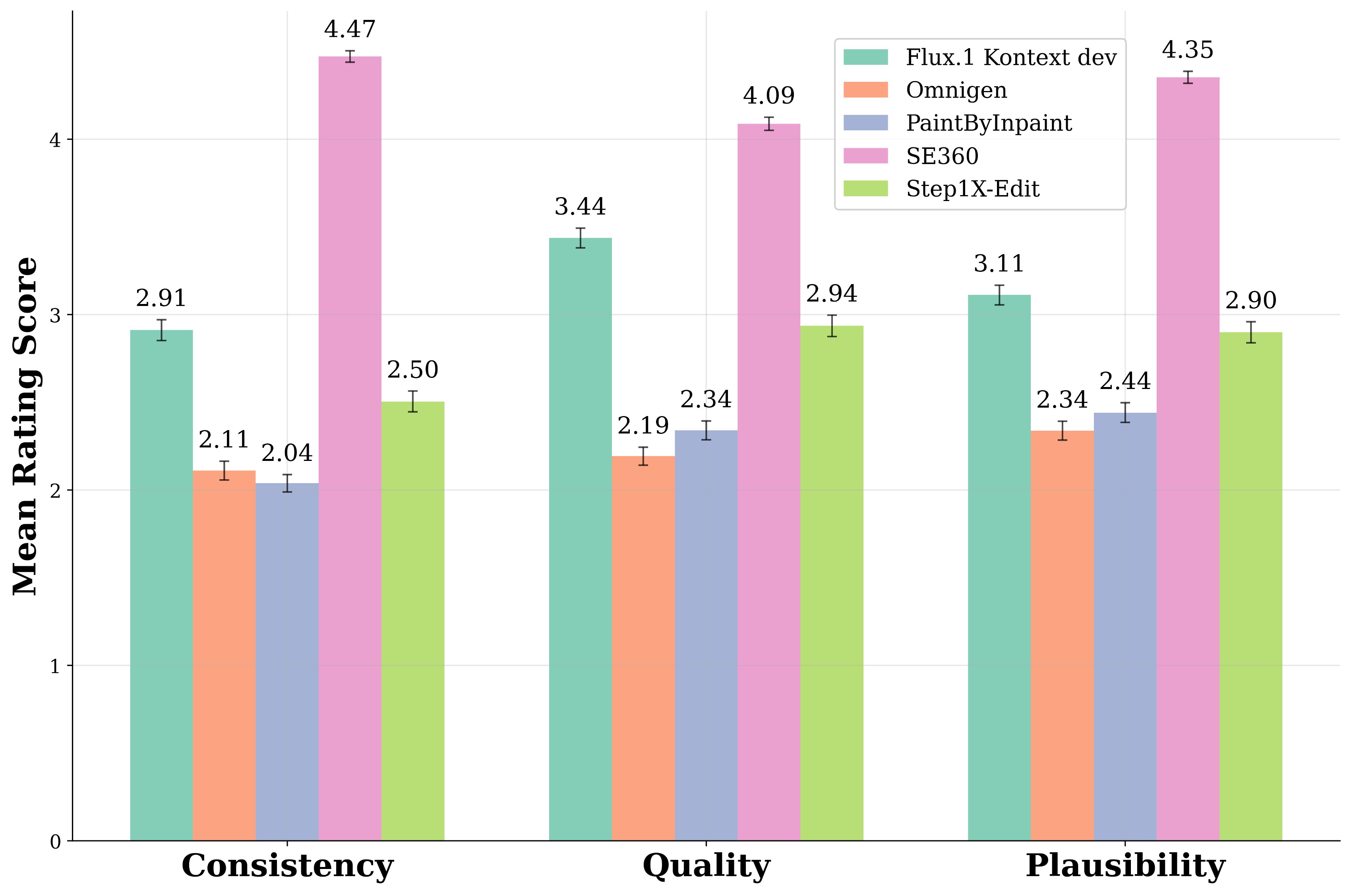} 
\caption{Model performance comparison across evaluation dimensions.}
\label{figs9}
\end{figure}

\begin{figure*}[t!]
    \centering
    \includegraphics[width=2.00\columnwidth]{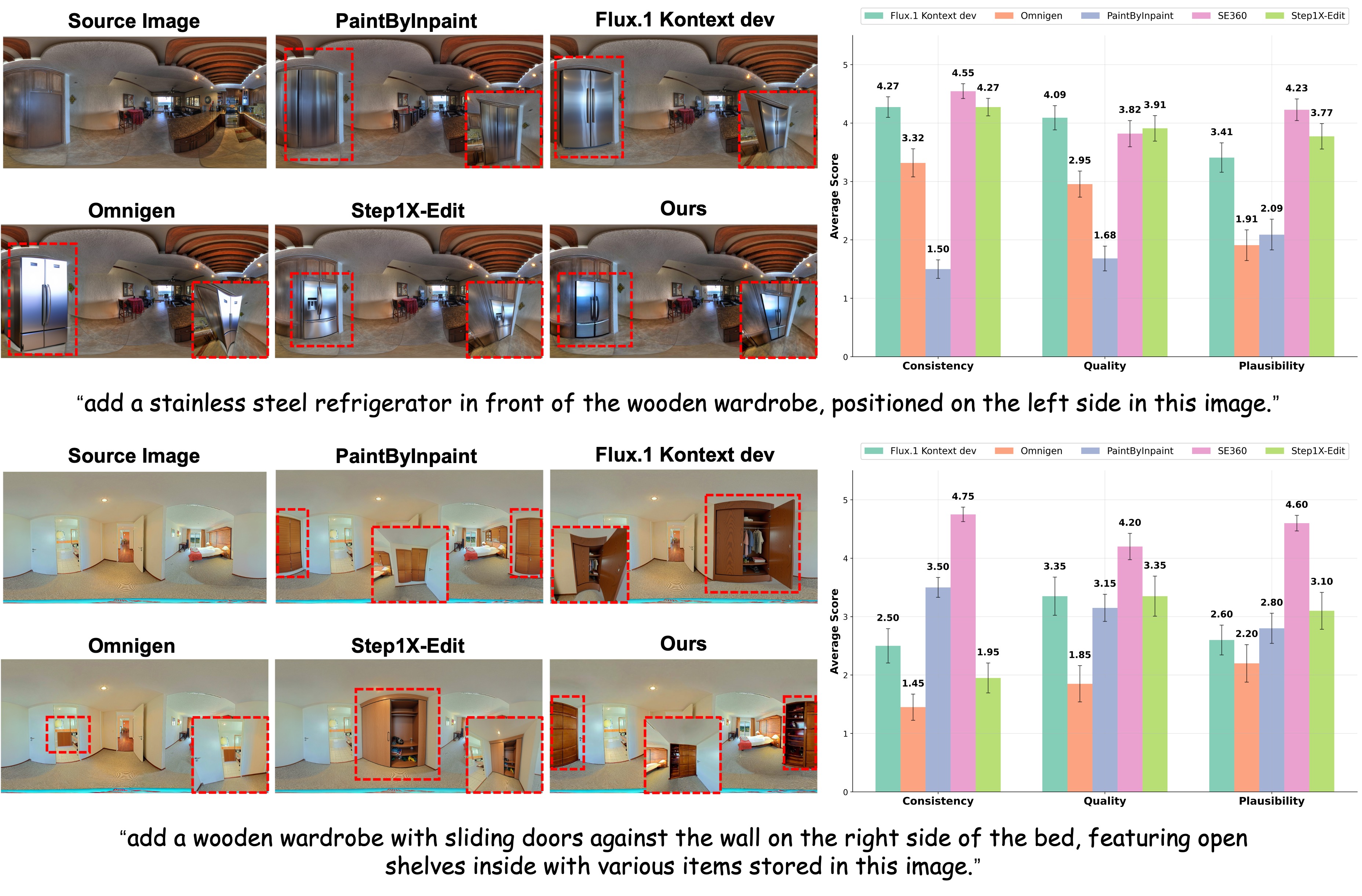}
    \caption{Two qualitative examples from our user study. Each case shows the generated results from five models, along with a corresponding bar chart of the average user scores. These examples visually demonstrate SE360's superiority in both generation quality and user preference.}
    \label{figs10}
\end{figure*}

These individual examples are representative of the overall trend. In addition to the mean scores reported in the main paper (see Table 4), we provide more detailed statistical charts in Figure~\ref{figs9} and Figure~\ref{figs12} to offer a comprehensive view of the user rating distributions.
\begin{itemize}
    \item \textbf{Bar Chart}: The bar chart (Figure~\ref{figs9}) visualizes the mean rating scores and their error margins for each model across the three criteria. This is consistent with the tabular data in the main paper and clearly shows that our method, SE360, significantly outperforms all other models on all metrics.
    \item \textbf{Box Plot}: The box plot (Figure~\ref{figs12}) further reveals the complete distribution of the scores, including the median (the line in the box), the interquartile range (the height of the box), the data range (the whiskers), and outliers (the diamond markers). From this plot, we can observe that the box for SE360 is not only positioned highest overall but is also relatively compact. This indicates that users consistently gave high scores with strong agreement. In contrast, other models have wider score distributions and more outliers, suggesting their performance is less stable and more controversial among users.
\end{itemize}
Together, these two charts demonstrate the overwhelming superiority of our method in terms of both generation quality and perceived user preference.

\begin{figure*}[t!]
\centering
\includegraphics[width=2.00\columnwidth]{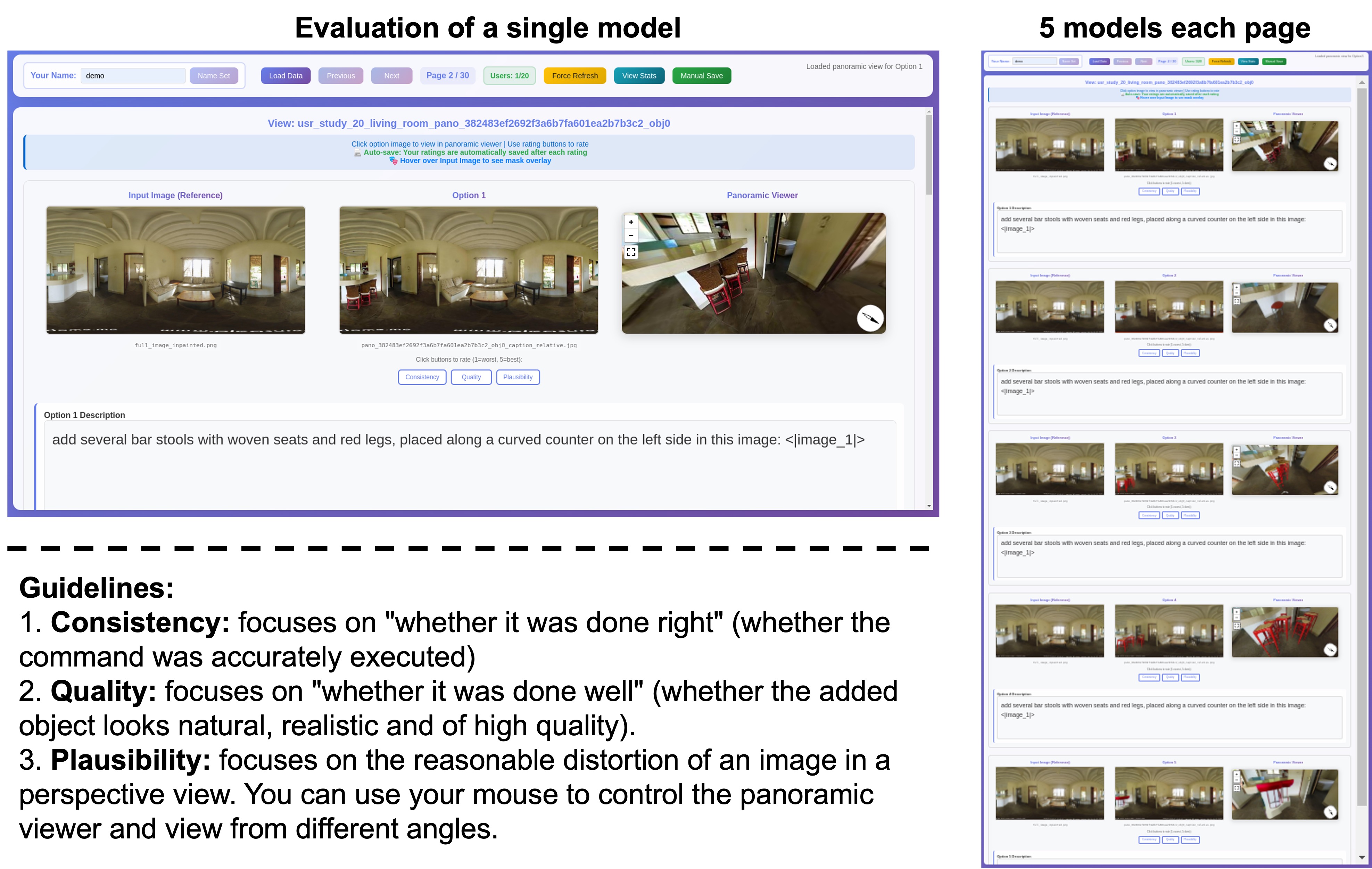}
\caption{Demonstration of user study page.}
\label{figs11}
\end{figure*}

\begin{figure*}[t!]
\centering
\includegraphics[width=2.00\columnwidth]{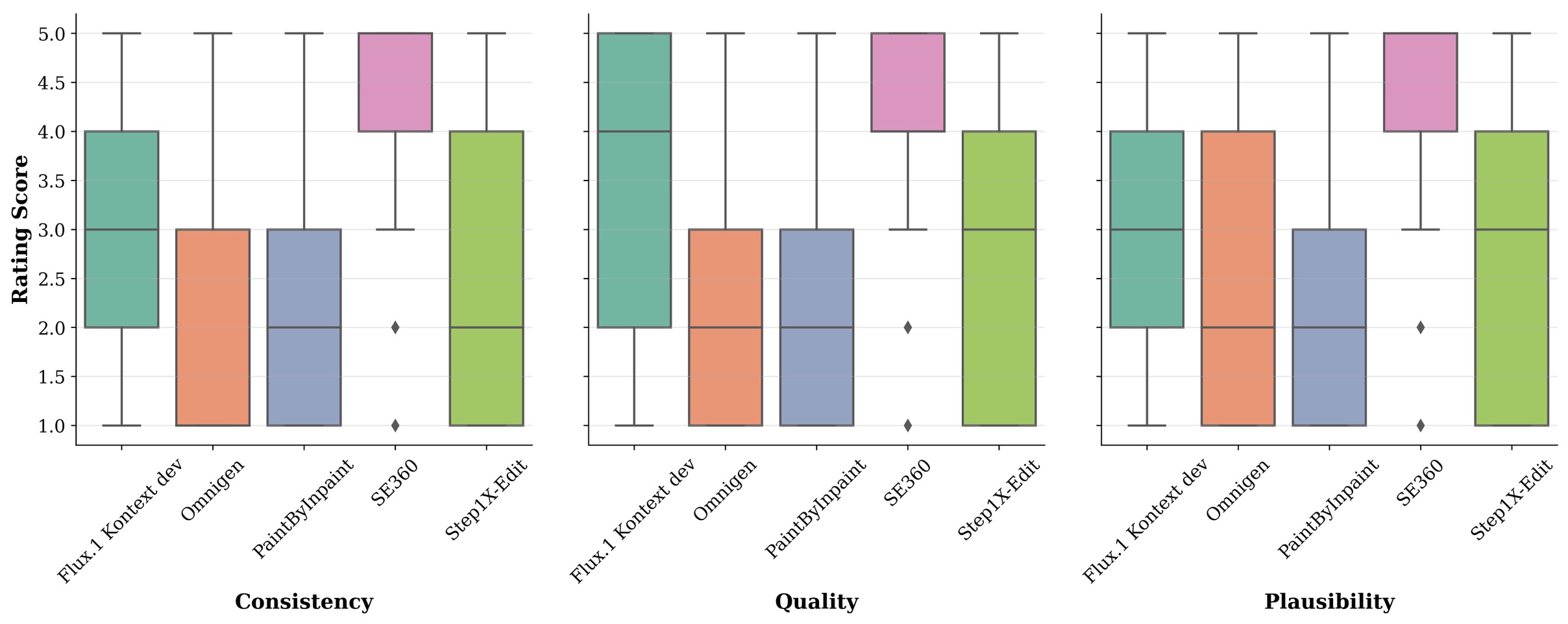}
\caption{Distribution of user ratings by model and evaluation dimension.}
\label{figs12}
\end{figure*}

\begin{listing*}[t]
\caption{VLM Prompt for Object Description in ERP Generation}
\label{lst:vlm_prompt_objcaption_generation}
\begin{lstlisting}[language=]
Describe the main objects in the foreground of the image. Don't include background or structure elements like walls, sky, stairs, ground, door, window, pole, railing, etc. List at most 8 objects.
         Prefer to choose completely visible objects, no occlusion. Don't choose objects that are close to the image edge.
         Please respond using the following fixed format for easier processing:

         ```json
         [
           {
             "description": "Detailed description of the object including color, shape and relative position. Only one sentence",
             "category": "Specific object name (be precise, not general category, only one name, don't use "/" and "or", 1-2 words)"
           },
         ]
         ```

         Notes:
         1. The category name should be the specific object name (e.g., "wooden coffee table", "ceramic vase", "leather sofa"), not just its general type
         2. Each description should include color, shape/type, and relative position (next to, behind, in front of, etc.).
         3. If you can't find 8 objects that meet the criteria, you can return fewer items, never choose room structure, even if you only find one object.
         4. Prioritize objects in the center of the image
         5. Don't choose the same objects multiple times.
         6. As some items are together, you can describe them as a group. For example, some books are arranged in a row; you can describe them as a group. Some chairs around a table, you can describe the table and chairs as a group.
         7. Don't include background or structure elements like walls, sky, stairs, ground, door, windows, pole, railing, etc.
\end{lstlisting}
\end{listing*}

\begin{listing*}[t]
\caption{VLM Prompt for Object Grounding in ERP}
\label{lst:vlm_prompt_objgrounding}
\begin{lstlisting}[language=]
Please locate the objects in the image based on the following descriptions and their associated categories. For each object found, return its bounding box coordinates (as a list [x1, y1, x2, y2]) along with the original description and category provided.

        Descriptions and Categories:
        {descriptions_categories_text}

        For each description-category pair provided above, find the corresponding object in the image and return its bounding box.

        ###
        # Required Output Format:
        The output MUST be a single JSON list enclosed in ```json ... ```.
        Each element in the list MUST be a JSON object.
        Each JSON object MUST contain EXACTLY THREE keys: "description", "category", and "bbox_2d".
        - The value for "description" MUST be one of the original descriptions provided in the input.
        - The value for "category" MUST be the category associated with that description in the input.
        - The value for "bbox_2d" MUST be a list containing EXACTLY FOUR numbers (integer or float): [x1, y1, x2, y2].
        NO OTHER FORMATS ARE ACCEPTABLE. Do NOT put coordinates in a string. Do NOT omit keys.

        Output format example:
        ```json
        [
         {{"description": "description_1", "category": "category_name_1", "bbox_2d": [x1, y1, x2, y2]}},
         {{"description": "description_2", "category": "category_name_2", "bbox_2d": [x1, y1, x2, y2]}},
         ...
        ]
        ```
        ###
        # Important Rules:
        - If you find multiple instances for a description, choose the larger one. The bbox needs contain all the objects in the description. Use the original description and category.
        - If you cannot confidently locate a bounding box for a specific description-category pair, OMIT that object entirely from the JSON list output. Do not include entries with null, empty, or incorrectly formatted values.
        - Ensure all coordinate values in the "bbox_2d" list are numbers.
        - Ensure the returned "description" and "category" exactly match the ones provided in the input list for the located object.
        - If the category of the description carries other objects, you need to output the category and object as a whole in a single bbox.
        ###

        Provide the JSON output containing the bounding boxes, descriptions, and categories for the located objects, strictly following the required format.
\end{lstlisting}
\end{listing*}

\begin{listing*}[t]
\caption{VLM Prompt for PAA}
\label{lst:vlm_prompt_PAA}
\begin{lstlisting}[language=]
The chosen item is: {chosen_item_description}

Your task is to identify the `chosen_item` and list ONLY the items that are **directly and physically supported** by it. This means items that are:
1.  **Resting directly ON TOP** of the chosen item.
2.  **Contained INSIDE** the chosen item (like flowers in a vase).
3.  **Directly attached TO** the chosen item (less common in indoor scenes, but possible).
**Crucially, DO NOT LIST items that are merely:**
* Behind the chosen item.
* In front of the chosen item.
* Next to the chosen item.
* Visible through a gap or underneath the chosen item (like the floor under a table).
Think: If the `chosen_item` were instantly removed, ONLY list the items that would fall *because* the chosen_item is gone. Ignore items that are simply nearby or in the background/foreground.
List each physically supported item (and the chosen item itself) separately, one per line.
### Example 1:
a scene with a table, a vase with flowers, and two books on the table.
When the chosen item is a table, the vase, runner, and books are directly ON the table. Output:
- table (1)
- vase with flowers (1)
- table runner (1)
- book (2)
When the chosen items are books. Output:
- book (2)
### Example 2:
a scene with a sofa, some pillows on the sofa.
When the chosen item is a sofa, the pillows are directly ON the sofa. Output:
- sofa (1)
- pillow (4)
When the chosen items are pillows, it rests ON the sofa but does not support other items. Output:
- pillow (4)
### Example 3:
a scene with a vase, some flowers in the vase.
When the chosen item is a vase or a vase with flowers, it contains the flowers. Output:
- vase (1)
- flowers (1)
When the chosen item is a flower, it is contained WITHIN the vase but supports nothing else. Output:
- flowers (1)
### The format of the output should be:
- chosen item (quantity)
- supported item1 (quantity)
- supported item2 (quantity)
- ...
or
- chosen item (quantity)
The name of the supported item (if there is one) just outputs the name, not the description like "supported item:".
If there is no supported item for the chosen item, just output the name of the chosen item. The rug and carpet are not special items; if the chosen item is rug or carpet, just output the name of the chosen item.
Always include the quantity of each item in parentheses (1) if singular or the exact number if multiple. The "()" only contains the quantity, not the description.
For supported objects with thin surfaces, such as table runner, paper, tablecloth, towels, etc., also need to detect if they are directly supported.
The description of the chosen item and the supported item should not contain location information.
IMPORTANT: ONLY output the item list in the format shown above. DO NOT include any explanations, reasoning, or additional text. DO NOT include any lines starting with "###" or containing "Explanation". Keep your response concise and limited strictly to the item list format.
\end{lstlisting}
\end{listing*}

\begin{listing*}[t]
\caption{VLM Prompt for Instruction Recaption}
\label{lst:vlm_prompt_instruction recaption}
\begin{lstlisting}[language=]
Analyze the provided perspective IMAGE focusing on the object identified as "{params['original_category']}".
Use the Original Description ONLY for its location context: "{params['original_description']}"

Your task is to generate TWO descriptions based on the IMAGE and the location context:
1.  **Standard Refined Description:**
    * Describe the visual appearance (color, shape, material, state, anything *on* it) based *strictly and ONLY* on what you see in the perspective IMAGE.
    * Completely DISREGARD the appearance details mentioned in the Original Description.
    * Extract the spatial location context (e.g., "on the wall", "next to the sofa", "above the table") *strictly and ONLY* from the Original Description. Do not guess the location from the image.
    * Combine the visual details from the IMAGE and the location context from the Original Description into ONE fluent sentence (around 10-25 words).
2.  **Simple Description:**
    * Generate an EXTREMELY simple description (max 5-7 words).
    * State only the main color + object category + basic location derived from the Original Description's context.
    * Examples of desired simple output: "A blue armchair by the window.", "A silver lamp on the desk.", "A wooden table in the center."

--- Output Format ---
Provide your response EXACTLY in this format, with each description on a new line:

Standard Description: [Your standard refined description here]
Simple Description: [Your extremely simple description here]

--- Examples based on Input ---
Input Category: wooden table; Input Original Description (for location): a wooden table surrounded by chairs, centrally located in the room.
Output:
Standard Description: A light brown wooden table with a white vase containing pink flowers on it, surrounded by chairs in the center of the room.
Simple Description: A wooden table in the center.

Input Category: blue armchair; Input Original Description (for location): a comfortable blue armchair situated by the window.
Output:
Standard Description: A blue fabric armchair with a yellow knitted blanket draped over its back, situated by the window.
Simple Description: A blue armchair by the window.

Input Category: metal desk lamp; Input Original Description (for location): a metal desk lamp positioned on the left side of the desk.
Output:
Standard Description: A sleek silver metal desk lamp with an adjustable arm, positioned on the left side of the desk.
Simple Description: A silver lamp on the desk.
--- End Examples ---
Now, generate the two descriptions for:
Input Category: "{params['original_category']}"
Input Original Description (for location): "{params['original_description']}"
Output ONLY using the specified format. No extra text before or after.
\end{lstlisting}
\end{listing*}

\begin{listing*}[t]
\caption{LLM Prompt for Caption Classification}
\label{lst:llm_prompt_caption classification}
\begin{lstlisting}[language=]
Given the following object description from a 360-degree image context:
"{description}"

Determine if the description primarily uses absolute positioning or relative positioning.
- Absolute positioning refers to the object's location only concerning the overall image frame or room structure (e.g., 'center of the image', 'left side of the room', 'right side of the room', 'right of the image', 'centerly in the image').
- Relative positioning refers to the object's location concerning other specific objects (e.g., 'next to the chair', 'on top of the table', 'behind the sofa', 'left of the sink', 'right of the side table').

For example:
- "A white toilet with a closed lid, situated to the right of the sink area, near a textured wall." -> "relative"
- "A white toilet with a closed lid, situated to the right of the room." -> "relative" 
- "A black sofa, situated to the left of the image." -> "absolute"
- "A yellow chair, situated to the left of the table." -> "relative"
- "A blue desk in the center of the image." -> "absolute"
- "A red vase, placed on the right of the table." -> "relative"
- "A large illuminated mirror with a rectangular frame, mounted above the sink, reflecting part of the room." -> "absolute" 
- "A white rectangular sink with a modern design, positioned centrally on a curved vanity counter." -> "relative"
- "A flat-screen television mounted on the wall to the left, displaying colorful graphics." -> "absolute"
- "A large abstract painting with warm tones, hanging on the wall to the right of the dining area." -> "relative"
Respond with only one word: 'absolute' or 'relative'.
\end{lstlisting}
\end{listing*}

\end{document}